\documentclass[journal]{IEEEtran} 
\IEEEoverridecommandlockouts
\usepackage{fancyhdr}
\usepackage{xcolor}
\usepackage{tablefootnote}
\usepackage[
backend=biber,
style=ieee,
sorting=none
]{biblatex}
\usepackage{amsmath,amssymb,amsfonts,mathtools}
\usepackage{algorithmic}
\usepackage{bm}
\usepackage{makecell}
\usepackage{graphicx}
\usepackage[utf8]{inputenc}
\usepackage[english]{babel}
\usepackage{tabularx,booktabs}
\newcolumntype{C}{>{\centering\arraybackslash}X} 
\usepackage{lipsum}
\usepackage{authblk}
\usepackage{textcomp}
\usepackage{units}
\DeclareMathOperator*{\minimize}{minimize}

\newcommand{\sign}{\text{sign}}

\usepackage{subcaption}
\usepackage[export]{adjustbox}
\usepackage[export]{adjustbox}

\usepackage[ruled, lined, linesnumbered, commentsnumbered, longend]{algorithm2e}
\SetKwInput{KwInput}{Input}                
\SetKwInput{KwOutput}{Output}
\newcommand\mymathbb[1]
  { {\rm I\mathchoice{\hspace*{-2pt}}{\hspace*{-2pt}}
    {\hspace*{-1.75pt}}{\hspace*{-1.7pt}}#1}   }

\newcommand{\N}{\mymathbb N}
\begin{document}
\title{An NMPC-ECBF Framework for Dynamic Motion Planning and Execution in Vision-based Human-Robot Collaboration\\
\thanks{Dianhao Zhang is with the School of Electrical Engineering, Nantong University, Nantong 226019, China, and also with the Centre for Intelligent Autonomous Manufacturing Systems (i-AMS), School of Electronics, Electrical Engineering and Computer Science, Queen’s University Belfast, Belfast BT9 5AG, U.K. E-mail: dzhang07@qub.ac.uk.\\
Mien Van, Pantelis Sopasakis and Seán McLooneare with the Centre for Intelligent Autonomous Manufacturing Systems (i-AMS), School of Electronics, Electrical Engineering and Computer Science, Queen’s University Belfast, Belfast BT9 5AG, U.K. E-mail: dzhang07, m.van, p.sopasakis, s.mcloone@qub.ac.uk.}}
\author{Dianhao Zhang, Mien Van, Pantelis Sopasakis, Seán McLoone\IEEEmembership{, Senior Member,~IEEE}\\
This work has been submitted to the IEEE for possible publication. Copyright may be transferred without notice, after which this version may no longer be accessible.}


\maketitle
\begin{abstract}
   To enable safe and effective human-robot collaboration (HRC) in smart manufacturing, seamless integration of sensing, cognition, and prediction into the robot controller is critical for real-time awareness, response, and communication inside a heterogeneous environment (robots, humans, and equipment). The proposed approach takes advantage of the prediction capabilities of nonlinear model predictive control (NMPC) to execute safe path planning based on feedback from a vision system. In order to satisfy the requirement of real-time path planning, an embedded solver based on a penalty method is applied. However, due to tight sampling times, NMPC solutions are approximate, and hence the safety of the system cannot be guaranteed. To address this we formulate a novel safety-critical paradigm with an exponential control barrier function (ECBF) used as a safety filter. We also designed a simple human-robot collaboration scenario using CoppeliaSim to evaluate the performance of the proposed controller and investigate whether integrating human pose prediction can help with safe and efficient collaboration. The robot uses OptiTrack cameras for perception and dynamically generates collision-free trajectories to the predicted target interactive position. Results for a number of different configurations confirm the efficiency of the proposed motion planning and execution framework, with a 23.2\% reduction in execution time achieved for the HRC task considered when compared to an implementation without human motion prediction.
\end{abstract}

\section{Introduction}
With the advancements in technologies of motion prediction and embedded path planning, accurate and online human-robot collaboration (HRC) has been rapidly developed in smart manufacturing \cite{nicora2021human, 9302892}.
%
%
The learning space for executing human-robot collaboration tasks with a flexible joint manipulator is quite large and the dynamics are, in general, nonlinear, time-varying, and complex. Safe and smooth HRC is still an open challenge particularly for autonomous systems navigating in shared spaces with humans (e.g. intra–logistic and service robotics) and in densely crowded environments \cite{triebel2016spencer, proia2021control}. For these scenarios, it is necessary to develop robots with the ability to understand the ongoing task, accurately detect the human's position, and infer upcoming steps based on human demonstrations \cite{wang2018facilitating}.

 Vision-based human action recognition and motion prediction are crucial problems in an HRC system \cite{TERRERAN2023104523}. Action recognition aims to classify the categories of a human's current dynamics. To activate a robot at the proper time, the estimated classification category can be used as a prior condition for motion prediction. Motion prediction is concerned with forecasting future body movements and poses based on observations of past movements, and can be used to improve the collaboration efficiency in HRC \cite{8360558}. Estimates of the future evolution of body poses can be used to define both the target position for the robot end-effector and the dynamic obstacles (body parts) for collision avoidance. In our work, an OptiTrack camera is adopted to achieve high-performance optical tracking combined with wearable devices in order to obtain a streaming skeletal representation of the human body. This data is then used as input to the task recognition and motion prediction modules.

Safety regulations ISO 10218 and the technical specification ISO 15066 \cite{villani2018survey} must be adhered to if a robot is deployed in a shared working environment with a human. These define cooperative robot operation as being contact type (e.g. power and force limiting (PFL), hand guiding (HG)) or non-contact type (e.g. safety-rated monitored stop (SRMS), speed and separation monitoring (SSM)) interactions. In this work, we consider the design of a control framework for SSM-based non-contact HRC, which guarantees safety by remaining a safe distance from humans during interactions.

We employ a model predictive control (MPC) based control system to meet the safety standards and process the online feedback from a vision system. Since the HRC environment is dynamic and stochastic, and cannot be fully predicted a priori,  MPC has become more and more popular in motion planning because it can handle various kinds of kinematic and dynamic constraints \cite{ide2011real, salaj2015pendubot}. The nonlinear model predictive control (NMPC) law is computed by solving a nonlinear optimal control problem online, taking into account a kinematic model. In \cite{li2019dynamical}, NMPC is adopted to solve the task-constrained motion planning problem. Considering safety in physical human-robot interaction (pHRI), \cite{oleinikov2021safety} proposes an online NMPC method using a kinematic model based on feedback from a vision system. 

NMPC can guarantee asymptotic stability and collision avoidance in most cases \cite{RawMayDie22}. However, infeasible solutions might be obtained in some cases, especially within the stringent sampling time requirements encountered in HRC (typically a few milliseconds).
%
%
%
This presents challenges for the real-time computation of NMPC as there is the potential for collisions if an inaccurate solution is computed within the available time \cite{wolf2016fast}. To overcome this limitation of NMPC, it is necessary to calculate the minimum distance between the end-effector and a human's interactive body component more accurately and to add a safety filter to constrain the robot's behavior within a safe area. To solve this problem, we use the Gilbert-Johnson-Keerth (GJK) algorithm \cite{secil2022minimum} to solve the collision detection problem and add a control barrier function (CBF) based constraint into the NMPC problem to guarantee the safety of the operator. In \cite{wang2022high}, a CBF-based approach is presented to constrain a redundant manipulator within a safe working area when interacting with a human operator. High-order CBF (HOCBF) and exponential CBF (ECBF) extensions are proposed in \cite{son2019safety, wang2021learning} to solve higher-order relative degree problems. In this paper, we combine a vision system with NMPC-ECBF to solve the safe pHRI problem online. 

In summary, the existing research gaps are as follows. Firstly, there is no established approach for designing a safe controller that uses the information from human motion prediction. Secondly, collaboration under a tight-sampling regime limits the accuracy of the NMPC solver, leading to errors that may present a risk to the human operators in the shared workspace.

With estimates of the future target and obstacle position in the working environment, we propose a real-time motion planning algorithm for the physical human-robot interaction (pHRI) task based on NMPC-ECBF. In this work, to test the performance of the proposed controller, we set up a HRC scenario with a 7 degrees of freedom (DOF) manipulator (a Baxter robot) whereby motion capture, action recognition, motion prediction, and trajectory planning modules are integrated into a single system. The scenario, a screw-driver usage task, is divided into a sequence of sub-tasks with the robot end-effector required to reach a continuously updated predicted interactive position without collision when a sub-task is triggered. However, as human motion and pose predictions are subject to approximation errors, the risk of collision still exists in HRC. Therefore, the output configuration of a manipulator is followed by a safety-critical controller implemented using an ECBF-based approach. 

There have been several works \cite{liu2017human} that consider integrating motion prediction into HRC tasks, but no corresponding safe controllers have been proposed to guarantee the safety of human operators. In this work, we not only perform path planning based on the predicted human motion but also develop a safety filter following the NMPC path planner to enhance the safety of human operators and ensure satisfaction with the ISO safety regulation. This makes the approach applicable to real-world manufacturing.

The main contributions of the paper are as follows:

(1) We design a controller consisting of a NMPC-based high-level controller and a safety-critical-control based low-level controller to satisfy the demand of a HRC task with frequently interactions between the human and the robot, which guarantees the safety of the human and reduces the idle time for agents in HRC compared with the existing HRC models in manufacturing \cite{liu2019deep}.

(2) Since the trajectory generated by NMPC has the potential for collision because the solver only solves the first-order optimality conditions and does not guarantee a global
optimum we employ an ECBF to enforce constraint satisfaction and collision avoidance while taking advantage of the path-planning properties of NMPC. 

(3) We develop a real-life human-robot collaboration scenario to test the safety and efficiency of the proposed human motion prediction enhanced HRC control methodology.

This paper is organized as follows: Section \ref{sec:overall} describes the overall HRC system. Related work is introduced in Section \ref{sec:relate}. The formulations and derivations of the target problem are presented in Section \ref{sec:problem}. Section \ref{sec:NMPC} discusses the methodology for solving the problem, consisting of formulations for the system, obstacles detection, introduction to the NMPC algorithm and the solver used in the optimization problem. The controller design is introduced in Section \ref{sec:controller}. Then, Section \ref{sec:ex} describes the experimental setup and presents the results of the simulations conducted to evaluate the performance of the proposed NMPC-ECBF based control solution for human-robot collaboration. Finally, Section \ref{sec:con} gives the conclusions.

\textbf{Notation:} The following notation is used throughout the paper. Let $\mathbb{R}$, $\mathbb{R}_{+}$, $\mathbb{R}^n$, $\mathbb{R}^{m\times n}$ and $\N$ denote the sets of real numbers, the sets of positive real numbers, the set of real vectors of length $n$,  the set of $m$ by $n$ real matrices and non-negative integers, respectively. Let $x$, $\bm{x}$, and $\bm{X}$ denote scalar, vector and matrix quantities, respectively, $\mathcal{X}$ a set, and $X$ a constant.
For any non-negative integers $k_1 < k_2$ the finite set $\{ k_1, \dots, k_2 \}$ is denoted by $\N_{[k_1, k_2]}$. For $\bm{x}\in \mathbb{R}^n$, $[\bm{x}]_{+} = \max \{0,\bm{x}\}$ is defined element wise. Given a set $C$, $Inf C$ gives the maximum lower bound of the set $C$, while $sup C$ gives the least upper bound of the set $C$.


\section{Overall System Design}\label{sec:overall}
The overall design of the collision-free HRC system is shown in Fig 1. The proposed HRC system consists of motion capture, action recognition, motion prediction, decision-making, path planning, and robot controller modules. The motion capture module relies on OptiTrack cameras to generate high-quality 3D streaming skeleton pose estimates at a sampling rate of 20 frames per second. Using recorded spatial-temporal data for a human performing task-related motions/activities as training data, a motion prediction module is pretrained to predict the future one-second evolution of human poses (20 frames) based on the motion observed over the previous one-second interval. The decision-making module is implemented using NMPC-ECBF based motion planning and robot execution mechanisms. In our work, a robot starts from a random position in the working environment and continuously searches for a collision-free path to the estimated interaction coordinates based on the received 1-second ahead pose estimation feedback provided by the predictor module.
\begin{figure}
\centering
\includegraphics[width=\columnwidth]{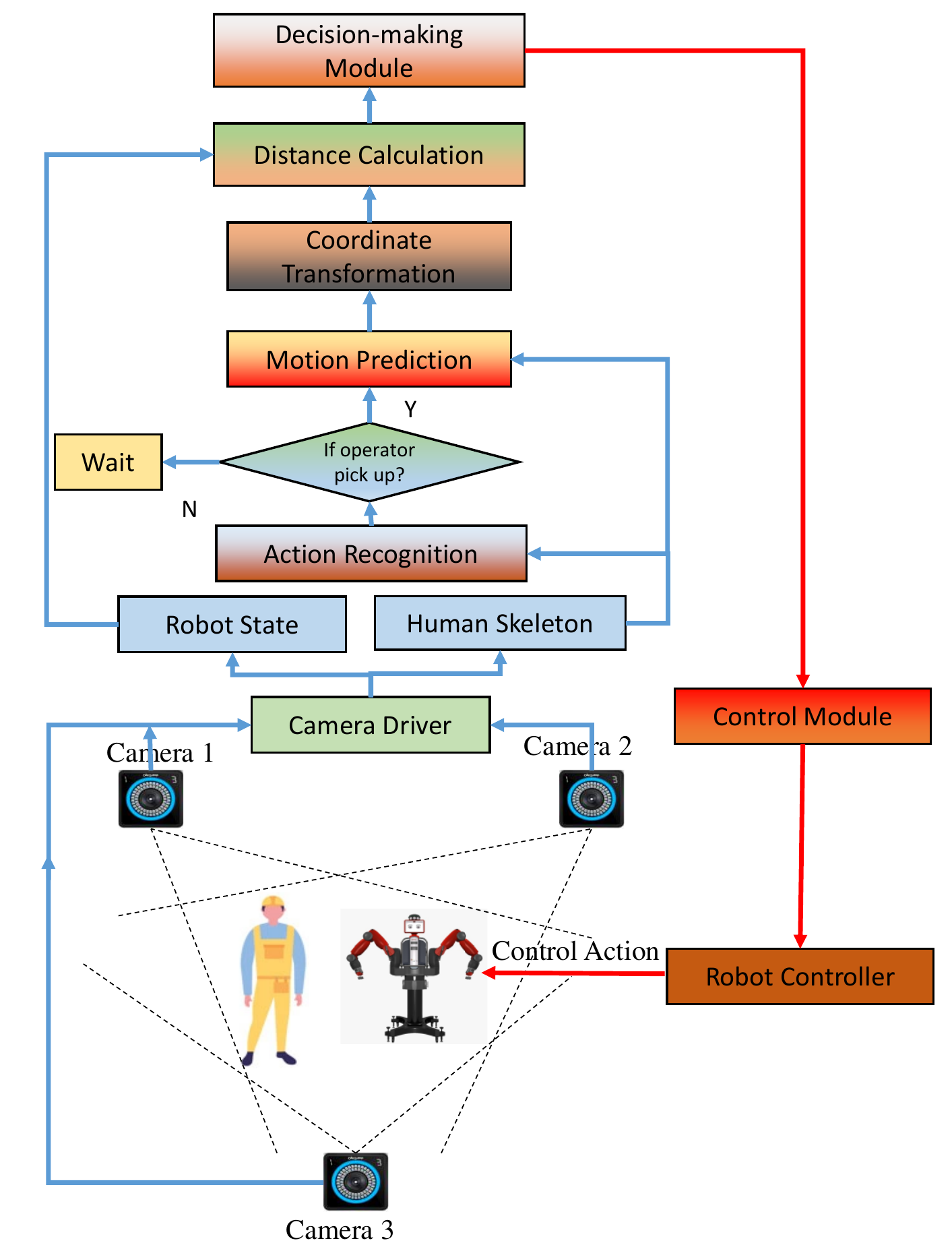}
\caption{Proposed human-robot collaboration system design}
\label{fig:overview}
\end{figure}
\begin{figure}[t]
\centering
\includegraphics[width=\columnwidth]{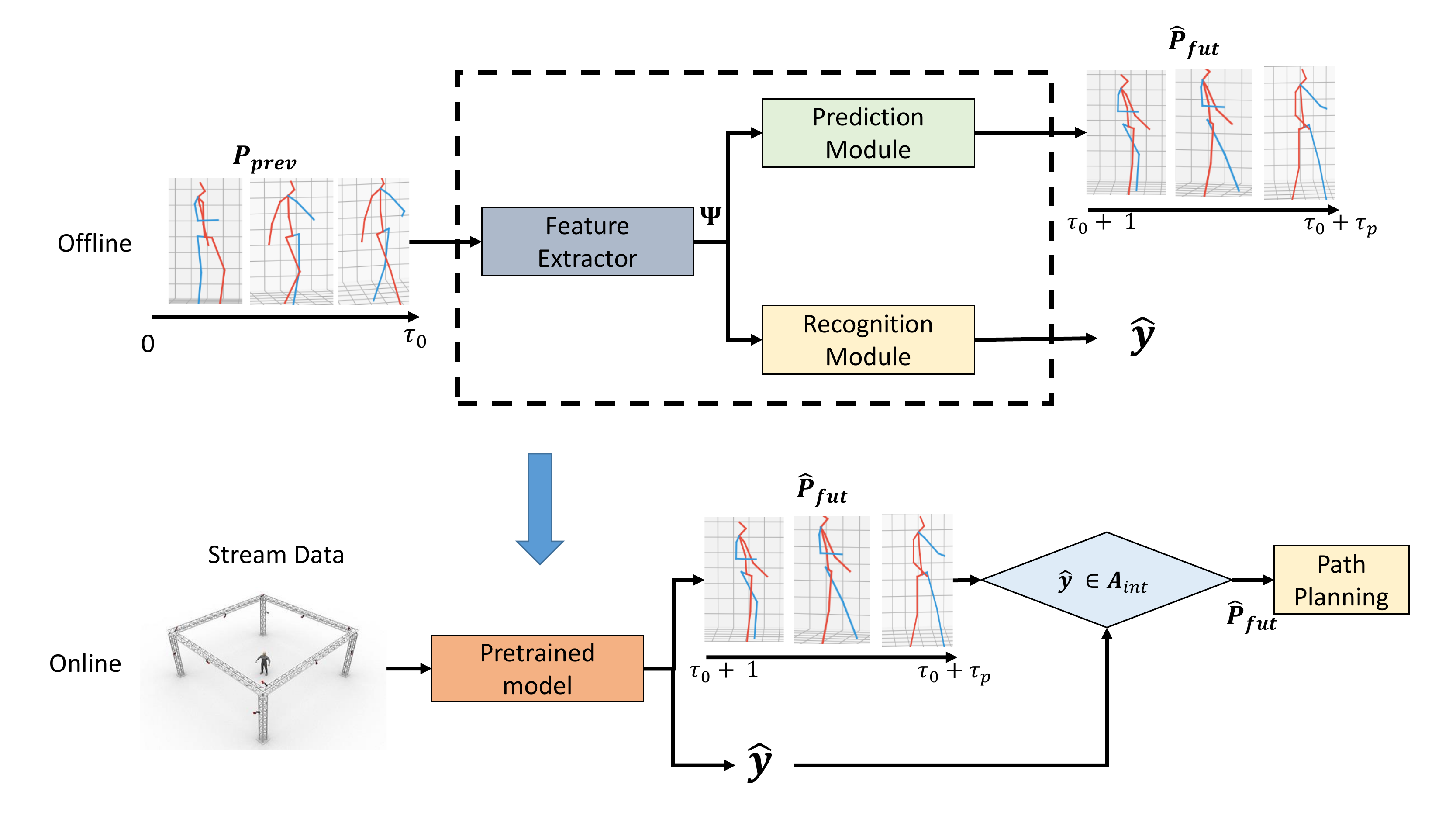}
\caption{Work flow from non-local graph convolution (NGC) based vision system to path planning. The outputs from the recognition and prediction modules are estimated human action category and future pose sequence, respectively.}
\label{fig:pretrain}
\end{figure}
\section{Related Works}\label{sec:relate}

\subsection{Methods to solve trajectory optimization in MPC}
When the system is linear, the constraints are affine and the cost functions are quadratic, the associated MPC optimization problem is a quadratic program. Several convex optimization algorithms have been proposed \cite{lindqvist2020nonlinear, lindqvist2020collision} which are fast, robust and possess global convergence guarantees. In contrast to the first-order approaches mentioned above, Newton-type methods are able to deal with general constraints and have faster rates of convergence \cite{jerez2014embedded}. This leads to a considerable reduction in computation time and makes them better suited to handling real-time nonlinear problems.

The proximal averaged Newton-type method for optimal control (PANOC) is applied for optimal control considering obstacle avoidance \cite{sathya2018embedded}. Compared with SQP and IP, PANOC does not require the solution of linear systems or quadratic programming problems at every iteration and it involves only very simple operations such as vector additions, and scalar and inner products. Additionally, PANOC is able to converge globally to a point that satisfies the first-order optimal conditions of the problem from any initial guess. As an extension, \cite{sopasakis2020open, trimble2020context} combines the proximal averaged Newton-type method for optimal control (PANOC) with the penalty and augmented Lagrangian methods to compute approximate stationary points of non-convex problems.  

Although MPC is widely used in practice, it is known that suboptimality and infeasibility due to the numerical optimization method can compromise the stability and positive invariance (satisfaction of constraints) properties of the MPC-controlled system. There have been a few works such as \cite{rubagotti2014} that attempt to remedy this issue, but these are limited to linear dynamical systems, convex formulations, and specific numerical optimization methods. To the best of the authors’ knowledge, no results exist on the problem of NMPC stability under inexact numerical optimization.
In this work, we focus on the issue of collision avoidance, and we demonstrate that the combination of nonlinear MPC, PANOC, and a CBF-based framework leads to a significant improvement in safety in terms of obstacle avoidance.

\subsection{Safety-critical Manipulator Control}
A system is commonly defined to be safe when its state never leaves some chosen set, known as the safety set. This forms the basis of controlled invariance \cite{gurriet2020scalable}, that is, finding a control strategy that ensures the system always remains in the safety set. In \cite{gurriet2020scalable}, a safety filter is used in the control of a nonlinear system to restrict the desired inputs in a way that ensures safety of the system when necessary. The method proposed in \cite{nguyen2021robust} formulates robust control Lyapunov and barrier functions to provide guarantees of stability and safety in the presence of model uncertainty. A control barrier function (CBF) is applied in our work to solve the collision avoidance problem. Applications of the CBF approach to the control of a redundant manipulator can be found in \cite{singletary2021safety, landi2019safety}.

\section{Problem Formulation}\label{sec:problem}
Formally, the skeleton-format input to the action recognition and motion prediction modules is represented by 3D joint positions of $m$ joints for a sequence of frames recorded over time, which is the same format as in \cite{zhang2021non}. The action pose at time (frame) $t$ is $\bm{p}(t) \in \mathbb{R}^{3m}$. The observed time length is $t \in [1, \tau_o]$ and the unobserved  time length is $t \in [\tau_o + 1, \tau_o+\tau_p]$ for prediction.  

For a single human action sequence belonging to a set of $N_a$ possible actions, $\mathcal{A}=\{1,\ldots,N_a\}$, we have the observed and future motion matrices, $\bm{P}_{\mathrm{prev}}$ and $\bm{P}_{\mathrm{fut}}$, where $\bm{P}_{\mathrm{prev}} = [\bm{p}(1), \cdots, \bm{p}(\tau_o)] \in \mathbb{R}^{\tau_o \times 3m}$ and $\bm{P}_{\mathrm{fut}} = [\hat{\bm{p}}(\tau_o + 1), \cdots, \hat{\bm{p}}(\tau_o+\tau_p)] \in \mathbb{R}^{\tau_p\times 3m}$ and the action classification label encoded as a one-hot vector $\bm{y} \in \{0,1\}^{N_{a}}$. The predicted human pose at time $t$ is denoted as $\hat{\bm{p}}(t)\in \mathbb{R}^{3m}$, which can be divided into the right hand's position $\hat{\bm{p}}_{rh}(t)\in \mathbb{R}^3$ as the target position for the robot and the remaining body joints' positions $\hat{\bm{p}}_{o}(t)\in \mathbb{R}^{3(m-1)}$ to decide the positions of obstacles. 

Human action recognition and motion prediction are achieved using the two-stage deep learning computer vision approach introduced in \cite{zhang2021non}, as depicted in Fig. \ref{fig:pretrain}. The 3D-skeleton data stream provided by the OptiTrack motion capture system is firstly processed by a bidirectional LSTM and a GCN architecture incorporating the attention mechanism to extract spatial-temporal features of the human motion dynamics. The generated feature map denoted as $\bm{\Psi}$, is then fed simultaneously to two modules, one that performs prediction and one that performs recognition. The motion prediction task is achieved using an LSTM-based decoder, while the action recognition task is performed using a CNN followed by conditional random fields (CRF). Initially, the models are trained offline using collected HRC task samples. Following training, streaming data is fed to the model, to generate the predicted future pose sequence $\hat{\bm{P}}_{\mathrm{fut}}$ and predicted action category $\hat{{y}}$. Denoting the feature extraction, motion prediction and action recognition functions within the Deep Neural Network model as $\bm{f}_{\mathrm{ext}}$, $\bm{f}_{\mathrm{pred}}$ and $\bm{f}_{\mathrm{recg}}$, respectively, and $\bm{\theta}_{\rm ext}$, $\bm{\theta}_{\rm pred}$ and $\bm{\theta}_{\mathrm{recg}}$ as their trainable parameters, the extracted feature map $\bm{\Psi}$, the estimated future motion $\hat{\bm{P}}_{\rm fut}$ and one-hot vector class category $\hat{\bm{y}}$ can be expressed as
\begin{subequations}
\begin{align}
\bm{\Psi} &= \bm{f}_{\mathrm{ext}}(\bm{P}_{\mathrm{prev}};\bm{\theta}_{\rm ext}),
\\
\hat{\bm{P}}_{\mathrm{fut}} &= \bm{f}_{\mathrm{pred}}(\bm{\Psi}; \bm{\theta}_{\mathrm{pred}}),
\\
\hat{\bm{y}}&=\bm{f}_{\mathrm{recg}}(\bm{\Psi};\bm{\theta}_{\mathrm{recg}}).
\end{align}
\end{subequations}

%
%

If the predicted action is one that requires robot interaction, the predicted final position of the right hand (denoted ${\hat{\bm{p}}}_{\mathrm{rh}}(\tau_p)$) is set as the target interactive position for the robot end-effector and the predicted final position of the other joints are set as the center of capsule-format obstacles.
%
%

The motion planning and execution control problem can then be stated as follows: Given the predicted future human motion sequence $\hat{\bm{P}}_{\mathrm{fut}}$ and the initial position of the robot arm, and assuming that: 
\begin{itemize}
\item  the dynamics of the manipulator are known; 
\item  the error in the prediction of the future human position is bounded; 
\item the maximum velocity of the human hand is slower than that of the manipulator; 
\end{itemize}
the objective is to determine a torque input $\bm{\tau}_{\rm act}$ to the robot, which enables the robot to track the predicted position of the human's right hand $\hat{\bm{p}}_{\mathrm{rh}}(t)$ to the target position and pose, while at all times avoiding collision with the human's body. 

%
%
%
Denoting the set of interactive human actions requiring robot interaction as $\mathcal{A}_{\rm int}$, where  $\mathcal{A}_{\rm int} \subset \mathcal{A}$,  the motion prediction and path planning tasks are triggered if the predicted HRC task $\hat{y} \in  \mathcal{A}_{\rm int}$, as shown in Fig. \ref{fig:overview}.
\begin{figure}
\centering
\includegraphics[width=0.8\columnwidth]{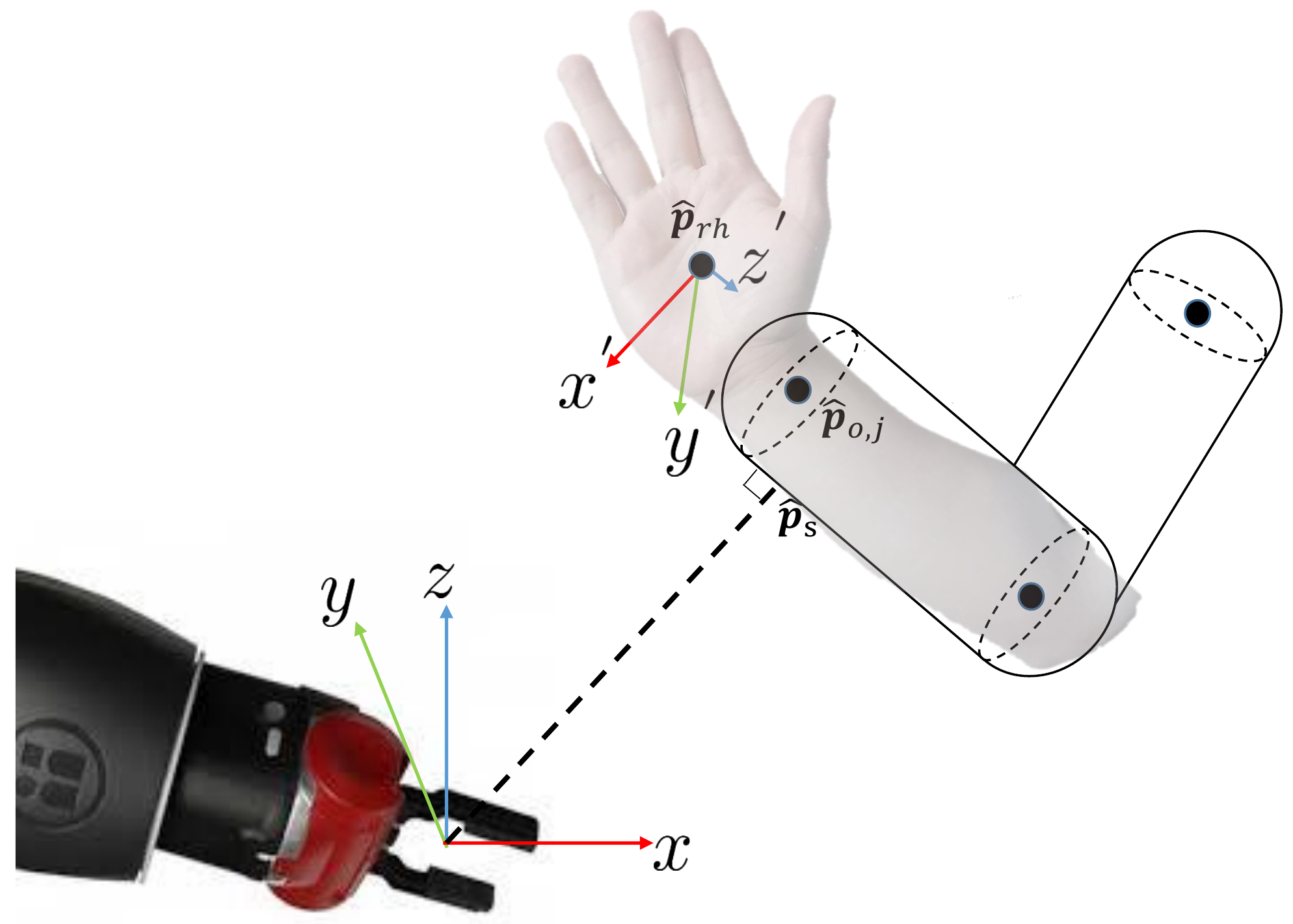}
\caption{Calculation of the distance, position and orientation of the capsules for the representation of humans.}
\label{fig:distance-calculation}
\end{figure}

\section{NMPC Motion Planning Scheme} \label{sec:NMPC}
\subsection{System Description}
To model the $n$-DOF robotic manipulator (chosen as the right arm of our Baxter robot) with joint positions $\bm{q}\in \mathbb{R}^n$, we consider an input-affine nonlinear system of the form
\begin{align}
\label{eq:affine}
    \dot{\bm{\chi}} &= \bm{\mathfrak{f}}(\bm{\chi}) + \bm{\mathfrak{g}}(\bm{\chi})\bm{\tau},
\end{align}
where $\bm{\chi} = [\bm{q}, \dot{\bm{q}}] \in \mathbb{R}^{2n}$ is the the state of the system and $\bm{\tau} \in \mathbb{R}^n$ is the control input. Functions $\bm{\mathfrak{f}}: \mathbb{R}^{2n} \rightarrow \mathbb{R}^{2n}$ and $\bm{\mathfrak{g}}: \mathbb{R}^{2n} \rightarrow \mathbb{R}^{2n\times n}$ are continuously differentiable in $\bm{\chi}$.

Consider an $n$-degrees-of-freedom manipulator with equations of motion defined as
\begin{align}
 \label{eq:dynamic}
     &\bm{M}(\bm{q})\Ddot{\bm{q}} + \bm{C}(\bm{q}, \dot{\bm{q}})\dot{\bm{q}} + \bm{g}(\bm{q}) = \bm{\tau},\\
     &\bm{x} = \bm{f}_{\mathrm{fwd}}(\bm{q}),
 \end{align}
where $\bm{M}(\bm{q}) \in \mathbb{R}^{n\times n}$ is a symmetric positive definite matrix, $\bm{C}(\bm{q}, \dot{\bm{q}}) \in \mathbb{R}^{n\times n}$ is a matrix of Coriolis and centrifugal forces terms, $\bm{g}(\bm{q})\in \mathbb{R}^n$ is a vector of gravitational forces, $\bm{q}\in \mathbb{R}^7$ are the joint positions, and $\bm{f}_{\mathrm{fwd}}: \mathbb{R}^7\rightarrow \mathbb{R}^3$ is the kinematic transformation
from the joint angle positions $\bm{q}$ to the Cartesian position of the end-effector $\bm{x} \in \mathbb{R}^3$. We consider the joint torque $\bm{\tau} \in \mathbb{R}^7$ as the system input and $\bm{x}$ as the system output.

Defining $\bm{J}=\bm{J}(\bm{q}) = \frac{\partial f_{fwd}}{\partial \bm{q}}$ as the Jacobian matrix it follows that
\begin{subequations}
\begin{align}
\label{eq:xdot}
    \dot{\bm{x}} ={}& \bm{J}(\bm{q})\dot{\bm{q}},\\
    \Ddot{\bm{x}} ={}& \bm{J}(\bm{q})\Ddot{\bm{q}} + \dot{\bm{J}}(\bm{q})\dot{\bm{q}}.
\end{align}
\end{subequations}
By substituting (\ref{eq:xdot}) into (\ref{eq:dynamic}), the Cartesian robotic system dynamics can be expressed as
\begin{align}
\label{eq:cartesiandy}
    \bm{M_x}(\bm{q})\Ddot{\bm{x}} + \bm{C_x}(\bm{q}, \dot{\bm{q}})\dot{\bm{x}} + \bm{g_x}(\bm{q}) = \bm{f_h},
\end{align}
where $\bm{M_x}(\bm{q}) = \bm{J}^{\dagger \intercal}\bm{M}(\bm{q})\bm{J}^{\dagger}$, $\bm{C_x}(\bm{q}, \dot{\bm{q}}) = \bm{J}^{\dagger \intercal}(\bm{C}(\bm{q}, \dot{\bm{q}})-\bm{M}(\bm{q})\bm{J}^{\dagger}\dot{\bm{J}})\bm{J}^{\dagger}$, $\bm{g_x}(\bm{q}) = \bm{J}^{\dagger\intercal}\bm{g}(\bm{q})$ and $\bm{f_h} = \bm{J}^{\dagger}\bm{\tau}$ is the vector of generated control forces. $\bm{J}^{\dagger}=\bm{J}^{\dagger}(\bm{q})$ denotes the pseudo-inverse of $\bm{J}(\bm{q})$, that is:
\begin{align}
    \bm{J}^{\dagger}(\bm{q}) = (\bm{J}^{\intercal}(\bm{q})\bm{J}(\bm{q}))^{-1}\bm{J}^{\intercal}(\bm{q}).
\end{align}
For notational simplicity, $\bm{M_x}$, $\bm{C_x}$, $\bm{g_x}$ will be used to denote $\bm{M_x}(\bm{q})$, $\bm{C_x}(\bm{q}, \dot{\bm{q}}$), $\bm{g_x}(\bm{q})$, respectively. The system (\ref{eq:affine}) is converted from joint space to task space, which can be expressed as 
\begin{align}
\label{eq:convdyna}
\begin{bmatrix}
\dot{\bm{x}}\\
\ddot{\bm{x}}
\end{bmatrix}=\underbracket[0.5pt]{\begin{bmatrix}
\dot{\bm{x}}\\
-\bm{M_x}^{-1}(\bm{C_x}\dot{\bm{x}}+\bm{g_x})
\end{bmatrix}}_{\bm{f}(\bm{x}, \dot{\bm{x}})}
+
\underbracket[0.5pt]{\begin{bmatrix}
0\\
-\bm{M_x}^{-1}
\end{bmatrix}}_{\bm{g}(\bm{x})}\bm{f_h}.
\end{align}
Thus, the robotic system (\ref{eq:convdyna}) is equivalent to the nonlinear affine system (\ref{eq:affine}) with respect to the input $\bm{f_h}$.  Therefore, the ECBF can be applied to the robotic system (\ref{eq:convdyna}) to deal with the time-varying output constraints.

\subsection{Minimum distance calculation}\label{Minimum distance calculation}
The minimum distance is the main input for most collision avoidance, HRI, robot decision-making, and robot navigation methods \cite{safeea2017minimum}. Given the joint coordinates provided by the skeletal tracking algorithm, bounding capsules (cylinders with semi-spherical ends) can be computed for the human body compo as a simplified representation of the human in the distance calculation. We generate minimum bounding capsules along the bones in the skeleton, as shown in Fig.\ref{fig:distance-calculation}. This is repeated for the predicted skeleton in each frame over the prediction horizon.  For the 32-joint skeleton model representation $\hat{\bm{p}}_{\rm\bm{rh}}$ is the position of the center of the right hand.

We choose the capsules model to represent robot links since the computational load to compute the distance is low \cite{lin2017real}. The 32-joint skeleton model, as used in the Human 3.6M dataset \cite{ionescu2013human3}, maps to a total of 15 capsules for the human body. The approach to measure the distance between capsules is depicted in Fig.\ref{fig:distance-calculation}.  The distance calculation is implemented using the GJK (Gilbert-Johnson-Keerthi) algorithm \cite{secil2022minimum}. 
The algorithm returns $\lambda(t)$ the minimum distance between the boundary of the capsules representing the human and the robot in the HRC task at time $t$.

\begin{figure}
\centering
\includegraphics[width=\columnwidth]{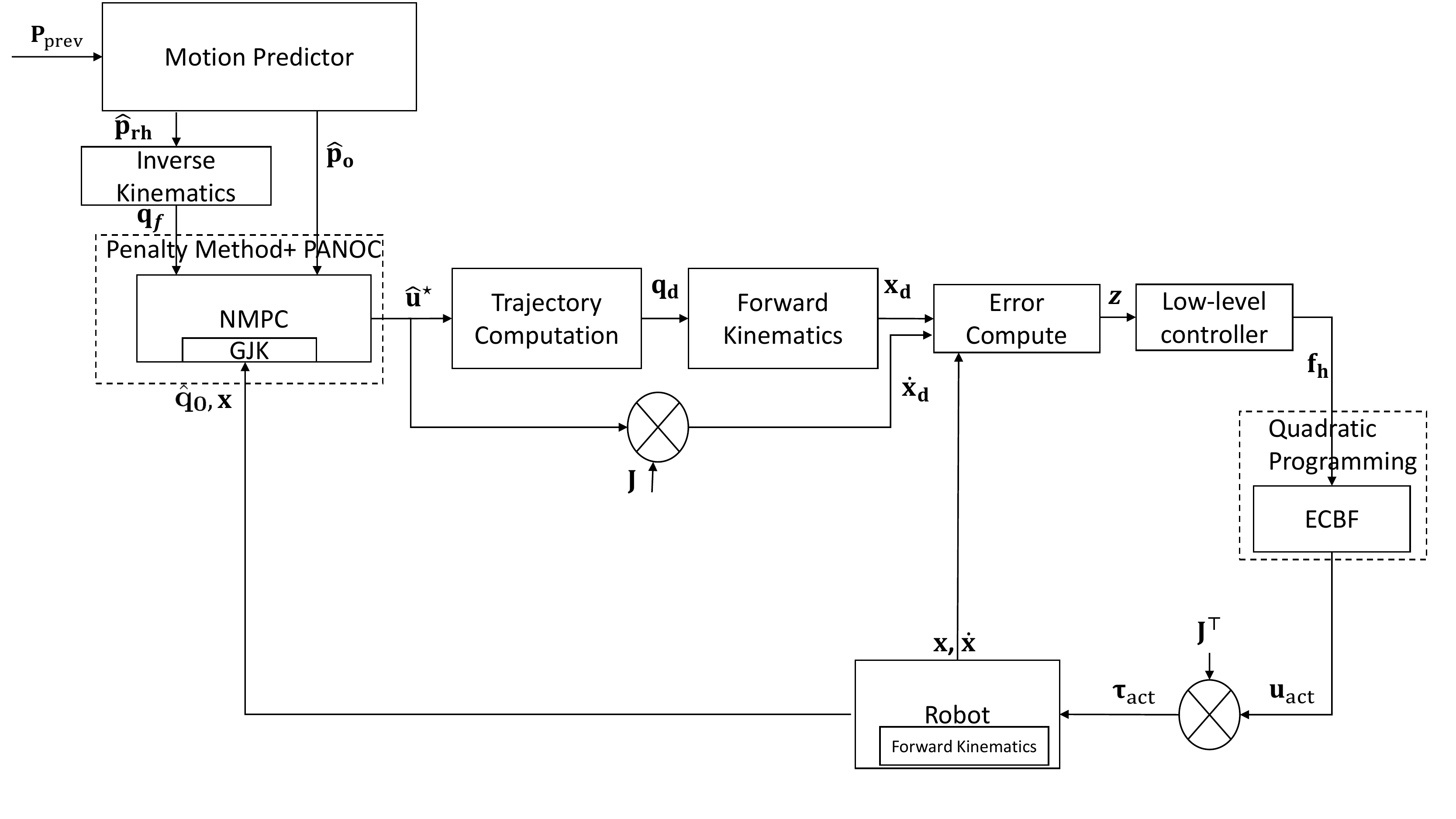}
\caption{NMPC-based motion planning system}
\label{fig:NMPC}
\end{figure}

\subsection{Nonlinear model predictive control for collision avoidance}\label{NMPC controller}
The robot’s objective is for its end-effector to follow the estimated position of the human's right hand while avoiding collisions. To that end, we propose an NMPC formulation as a high-level control system that commands angular velocity references to a low-level control system (see Section \ref{sec:torque-controller}). The overall controller architecture is shown in Fig. \ref{fig:NMPC}. 

The input of the high-level control system is the joint velocity commands $\bm{u}(t)$.
At every sampling time instant we need to solve a finite-horizon optimal control problem with prediction horizon $T_p > 0$, stage cost function $\ell$ and terminal cost function $\ell_f$,  where $\ell$ and $\ell_f$ are given by
%
\begin{equation}
\label{eq:stage}
\begin{split}
    \ell(\hat{\bm{x}},\hat{\bm{u}}) &= e(\hat{\bm{p}}_{\rm rh}, \hat{\bm{x}})^{\intercal}\bm{R}_p e(\hat{\bm{p}}_{\rm rh}, \hat{\bm{x}}) + \hat{\bm{u}}^{\intercal} \bm{R}_v \hat{\bm{u}},
    \end{split}
\end{equation}
and
\begin{multline}
    \ell_f(\hat{\bm{x}}) = \bm{e}(\hat{\bm{p}}_{\rm rh}(T_p), \hat{\bm{x}}(T_p))^{\intercal} \bm{Q}_p \bm{e}(\hat{\bm{p}}_{\rm rh}(T_p), \hat{\bm{x}}(T_p)) 
    \\
    + \hat{\bm{u}}^{\intercal}(T_p)\bm{Q}_v \hat{\bm{u}}(T_p),
\end{multline}
respectively. Here, $\bm{e}(\hat{\bm{p}}_{rh}(T_p), \bm{x}(t))$ is an error metric between the given Cartesian goal $\hat{\bm{p}}_{rh}(T_p)$ and current Cartesian state of the robot $\bm{x}(t)$. The reference position $\hat{\bm{p}}_{\mathrm{rh}}(T_p)$ is obtained by the motion predictor. $\bm{Q}_p$ and $\bm{Q}_v$ are constant weight matrices for terms in the terminal cost function, while $\bm{R}_p$ and $\bm{R}_v$ are the weight matrices for terms in the stage cost function. Following \cite{hu2021nmpc}, the error function $e(\bm{x}_d, \hat{\bm{x}}(t))$ is defined as
\begin{align}
    \bm{e}(\hat{\bm{p}}_{\rm rh}(t), \bm{x}(t)) =
    \begin{bmatrix}
    -\bm{I}_3 & 0\\
    0 & \bm{\Pi}_d
    \end{bmatrix}
    \hat{\bm{x}}(t) + 
    \begin{bmatrix}
    \hat{\bm{p}}_{\rm rh}(t)\\
    0
    \end{bmatrix},
\end{align}
where matrix $\bm{\Pi}_d$ is built from the desired quaternion, which remains constant during the optimization.

The initial and final configurations of the robot $\bm{q}_0 = [q_{01},\dots,q_{07}]$ and $\bm{q}_f = [q_{f1},\dots,q_{f7}]$ are obtained using the inverse kinematics function $\bm{f}_{\mathrm{inv}}: \mathbb{R}^6\rightarrow \mathbb{R}^7$, which converts the Cartesian end effector poses (position and orientation) to the corresponding joint positions. Using the GJK algorithm, the minimum distance calculation at time $t$ is performed based on the Cartesian coordinates of the robot, $\bm{x}(t)$, and the predicted human pose, $\hat{\bm{p}}_o(t)$. The distance between the end-effector and the predicted operator's position $\lambda(\hat{\bm{x}}(t); \hat{\bm{p}}_o(t))$ is required to be greater than the minimum safe distance $d_{\mathrm{safe}}$. The value of $d_{\mathrm{safe}}$ is chosen to be greater than the maximum prediction error bound to guarantee the safety of humans. The constraints are described as
\begin{subequations}
\label{eq:robot_constraints}
\begin{align}
&\dot{\hat{\bm{q}}}(t) = \hat{\bm{u}}(t),  t\in[0, T_p],
\\
&\hat{\bm{q}}(0) {}={} \bm{q}_0,
\\
&\hat{\bm{x}}(t) = \bm{f}_{\mathrm{fwd}}(\hat{\bm{q}}(t)),  t \in [0, T_p],
\\
&\lambda(\hat{\bm{x}}(t); \hat{\bm{p}}_o(t)) \geq d_{\mathrm{safe}}, t\in [0, T_p].
\end{align}
\end{subequations}
where $\bm{q} = (q_1, \dots, q_7)$ denotes the state of the robotic arm. The control actions are constrained in the set $\mathcal{U}$. 
Then, the optimal solution $\hat{\bm{u}}^{\star}$ is obtained by minimizing the optimal control cost function, that is,
\begin{equation}
\label{continuous}
    \minimize_{\hat{\bm{u}}:[0, T_p]\rightarrow \mathcal{U}} \ell_f(\hat{\bm{x}}(T_p))+ \int_{0}^{T_p} \ell(\hat{\bm{x}}(t), \hat{\bm{u}}(t))\mathrm{d}t,
\end{equation}
subject to the constraints (\ref{eq:robot_constraints}).

In this work, we solve the NMPC problem using a numerical optimization method. To that end, we convert the above problem to a discrete-time optimal control problem using an explicit discretization and integration method. Assuming that $T_p = N_h T_s$, the NMPC problem in Equation \eqref{continuous} is reformulated to $\mathbb{P}_d$ and depends on the current state $\bm{q}_0$, the set-point $\bm{q_f}$ for the joint positions, input $\bm{u}$ and the estimated obstacle distance $\hat{\lambda}$ computed by the GJK algorithm as follows
%
%
%
%
\begin{subequations}
\begin{align}
\label{eq:desprob}
    \mathbb{P}_d(\hat{\bm{q}}_0, \hat{\bm{p}}_{o}, \bm{q}_f):
    &\minimize_{{\hat{\bm{u}}_0,\dots,\hat{\bm{u}}_{N_h-1}}\in \mathcal{U}} 
    \ell_{d,f}(\hat{\bm{x}}_{N_h})\notag
    \\
    &\qquad\qquad {}+{} \sum_{k=0}^{N_h-1} \ell_d(\hat{\bm{x}}_k,  \hat{\bm{u}}_k),
\intertext{subject to the constraints}
\hat{\bm{q}}_{k+1} {}={}& \hat{\bm{q}}_k + T_s\hat{\bm{u}}_k, k\in \N_{[0,N_h-1]},\label{eq:step}
\\
\hat{\bm{q}}_0 {}={}& \bm{q}_0,
\\
\hat{\bm{q}}_{N_h} {}={}& \bm{q}_f,\label{eq:mpc:terminal}
\\
\hat{\bm{x}}_k {}={}& \bm{f}_{\mathrm{fwd}}(\hat{\bm{q}}_k), k\in \mathbb{N}_{[0,N_h-1]},
\\
\lambda(\hat{\bm{x}}_k; \hat{\bm{p}}_{o,k})  {}\geq{}& d_{\mathrm{safe}}, k\in \mathbb{N}_{[0,N_h-1]}.
\label{eq:mpc:collision-avoidance}
\end{align}
\end{subequations}
 By solving $\mathbb{P}_d$ at every discrete time instant, we obtain the solution $\hat{\bm{u}}_k^\star$, $k\in \mathbb{N}_{[0, N_h-1]}$, only the first value of which, $\hat{\bm{u}}_0^\star$, is applied to the system and $\mathbb{P}_d$ is solved again at the subsequent time instant in a receding horizon fashion.  

\subsection{Numerical solution}
To solve $\mathbb{P}_d$ numerically we employ OpEn \cite{sopasakis2020open}, a code generation software package that generates optimizers in the Rust programming language for deployment on embedded devices. OpEn combines the PANOC numerical algorithm with the penalty method to solve the above problem. However, OpEn requires that we cast $\mathbb{P}_d$ in the following general form 
\begin{subequations}
\begin{align}
    \minimize_{\bm{u} \in \tilde{\mathcal{U}}}\ &F_0(\bm{u}),
    \\
    \text{subject to: }
    &
    \bm{F}_i(\bm{u}) = \bm{0}, i\in\mathbb{N}_{[1, n_c]},
\end{align}
\end{subequations}
where $\bm{u}\in\mathbb{R}^n$ is the decision variable, $\tilde{\mathcal{U}} \subseteq \mathbb{R}^n$ is a nonempty closed set on which it is easy to compute projections, $F_0:\mathbb{R}^n\to\mathbb{R}$ is a continuously differentiable cost function --- not necessarily convex --- with Lipschitz gradient, and $\bm{F}_i:\mathbb{R}^n\to\mathbb{R}^{m_i}$ are functions such that $\|\bm{F}_i\|^2$ is continuously differentiable with Lipschitz gradient.

To cast problem $\mathbb{P}_d$ in this form, we choose the decision variable to be $\bm{u} = (\hat{\bm{u}}_0, \hat{\bm{u}}_1, \ldots, \hat{\bm{u}}_{N-1})$ and define the sequence of functions 
\begin{equation}
    \bm{\Phi}_{k+1}(\bm{u}, \bm{x}) {}={} \begin{bmatrix}
    \bm{q}_0 + T_s\sum_{t=0}^{k}\bm{u}_k
    \\
    \bm{f}_{\rm fwd}\left(\bm{q}_0 + T_s\sum_{t=0}^{k}\bm{u}_k\right)
    \end{bmatrix}
\end{equation}
with $\bm{\Phi}_0(\bm{u}, \bm{x}) = [\bm{q}^{\intercal}~~\bm{x}^{\intercal}]^{\intercal}$. Note that $\bm{\Phi}_{k+1}(\bm{u}, \bm{x})$ contains the location of the robot in both joint space and task space at time $k+1$ as a function of the initial state and the sequence of control actions $\bm{u}$. This allows us to eliminate the sequence of states and define the cost function 
\begin{equation}
    F_0(\bm{u}, \bm{x})
    {}={}
     \ell_{d,f}(\bm{\Phi}_{N_h}(\bm{u}, \bm{x}))
    {+} 
    \hspace{-0.5em}
    \sum_{k=0}^{N_h-1} \ell_d(\bm{\Phi}_{k}(\bm{u}, \bm{x}),  \bm{u}_k).
\end{equation}
In order to impose the terminal condition of Equation \eqref{eq:mpc:terminal} we define 
\begin{equation}
    \bm{F}_1(\bm{u}) = \hat{\bm{q}}_{N_h} - \bm{q}_f,
\end{equation}
where $\hat{\bm{q}}_{N_h} = \hat{\bm{q}}_{N_h}(\bm{u}, \bm{x})$. To impose the obstacle avoidance condition of Equation \eqref{eq:mpc:collision-avoidance} we define 
\begin{equation}
    \bm{F}_{k+1}(\bm{u}) = [d_{\rm safe} - \lambda(\bm{\Phi}_k(\bm{u}, \bm{x}); \hat{\bm{p}}_{o,k})]_{+},
\end{equation}
for $k\in\mathbb{N}_{[0, N_h - 1]}$. Lastly, we choose $\tilde{\mathcal{U}} = \mathcal{U}^{N_h}$ for the input constraints.

OpEn solves a series of optimization problems of the form 
\begin{equation}
    \mathbb{P}_{\rm in}:
    \minimize_{\bm{u}\in U}
    F_0(\bm{u}) + c \underbracket[0.5pt]{\sum_{i=1}^{n_c}\|\bm{F}_i(\bm{u})\|^2}_{\text{infeasibility}},
    \label{eq:inner-problem}
\end{equation}
for some $c>0$ termed the \textit{inner} problems. Essentially, the hard constraints of the original problem are turned into soft constraints. Starting with an initial value of $c$, a sequence of inner problems is solved numerically using the PANOC method while driving $c$ to infinity and warm-starting each next inner problem with the solution of the previous one \cite{sopasakis2020open}. The algorithm terminates once the infeasibility, as defined in Equation \eqref{eq:inner-problem} drops below a given tolerance $\epsilon_{\rm infeas} > 0$.

\section{Controller Design}\label{sec:controller}
\subsection{Trajectory Computation}
Given the estimated angular velocity command $\hat{\bm{u}}^{\star}$, the desired joint state $\bm{q}_d$ is updated as
\begin{align}
\label{eq:qdNMPC}
    \bm{q}_{d,k+1} = \bm{q}_{d,k} + T_s \hat{\bm{u}}^{\star}.
\end{align}
Then, the generated desired joint position is used to calculate the desired Cartesian position $\bm{x}_d$ and velocity $\dot{\bm{x}}_d$.

\subsection{Low-level controller}\label{sec:torque-controller}
In order to achieve asymptotic stability, the control force $\bm{f_h}$ input to the safety filter is defined as
\begin{multline}
\label{eq:controlforce}
    \bm{f_h} =\bm{C_x}(\dot{\bm{x}}_d - \Lambda \tilde{\bm{x}}) + \bm{g_x}+\bm{M_x}(\ddot{\bm{x}}_d - \Lambda\dot{\tilde{\bm{x}}})\\
    -k_z\sign(\bm{z}),
\end{multline}
where $\Lambda$ and $k_z$ are positive constants. A composite state error is denoted as $\bm{z} = \dot{\tilde{\bm{x}}} + \Lambda \tilde{\bm{x}}$, where $\tilde{\bm{x}} = \bm{x} - \bm{x}_d$ and $\dot{\tilde{\bm{x}}} = \dot{\bm{x}} - \dot{\bm{x}}_d$ are the error of the Cartesian position and velocity, respectively. $\sign(\bm{z})$ is a sign function depends on the state error. Considering the chattering problem caused by the $\sign$ function, \eqref{eq:controlforce} is rewritten as
\begin{multline}
\label{eq:controlforcenew}
    \bm{f_h} =\bm{C_x}(\dot{\bm{x}}_d - \Lambda \tilde{\bm{x}}) + \bm{g_x} + \bm{M_x}(\ddot{\bm{x}}_d - \Lambda\dot{\tilde{\bm{x}}})
    \\
    - k_z\frac{\bm{z}}{\Vert \bm{z} \Vert + c_1},
\end{multline}
where $c_1$ is a small positive number.

\subsection{Exponential Control Barrier Function as a Safety Filter}\label{Control Barrier Function}
In this section, an ECBF-based safety filter is introduced to prevent collisions with the human operator by constraining the trajectory generated by NMPC to be in a safe zone. ECBFs are formally defined as follows:
\\

\noindent\textbf{Definition 1 (ECBF \cite{nguyen2016exponential}).}
\noindent Consider the system (\ref{eq:convdyna}) and a set $\mathcal{C} = \{\bm{x} \in \mathbb{R}^3 | h(t,\bm{x})\geq 0, \forall t>0\}$ defined by a continuously differentiable function $h(t, \bm{x}):\mathbb{R}^3 \rightarrow \mathbb{R}$ with relative degree $\kappa \geq 2$. Then $h(t,\bm{x})$ is an exponential control barrier function (ECBF) if there is a $\bm{k}_b \in \mathbb{R}^{\kappa-1}$ such that
\begin{align}
    \inf_{\bm{u} \in \mathcal{U}}[L_{\bm{f}}^{\kappa} h(t, \bm{x}) + L_{\bm{g}}L_{\bm{f}}^{\kappa-1} h(t, \bm{x}) \bm{u} + \bm{k}_b \bm{\zeta_b}(\bm{x}) \geq 0,
\end{align}
where $\bm{\zeta_b}$ can be written as
\begin{align}
    \bm{\zeta}_b(\bm{x}):&=\begin{bmatrix}
        h(\bm{x})\\
        \dot{h}(\bm{x})\\
        \vdots\\
        h^{(\kappa-1)}(\bm{x})
    \end{bmatrix}
    =\begin{bmatrix}
        h(\bm{x})\\
        L_{\bm{f}} h(\bm{x})\\
        \vdots\\
        L_{\bm{f}}^{(\kappa-1)}h(\bm{x})
    \end{bmatrix}.
\end{align}
The method for selecting $\bm{k}_b$ is detailed in \cite{nguyen2016exponential}. The Lie derivative of $h(\bm{x})$ along $f(\bm{x})$ is defined as
\begin{align}
    L_{\bm{f}} h(t, \bm{x}) = \frac{\partial h(t, \bm{x})}{\partial \bm{x}} \bm{f}(\bm{x}, \dot{\bm{x}}),
\end{align}
where $\bm{x}$ is the state of the system. The Lie derivative of $h(\bm{x})$ along $\bm{g}(\bm{x})$ is defined as $L_{\bm{g}} h(t,\bm{x})$.
Denote $\bm{p}_{oi}(t) \in \mathbb{R}^3$ as the support point (i.e. a point on the boundary of the convex set that is closest to the robot) on the convex set representing the area of the human obstacle for the $i^{th}$ robot joint computed by GJK at time $t$. To ensure the robot does not collide with the moving human, the Cartesian position of the robot joints $\bm{x}_i \in \mathbb{R}^3$, $i\in \mathbb{N}_{[1, 7]}$ are under the following constraints
\begin{align}
\label{eq:safedistance}
    \Vert \bm{x}_i(t) - \hat{\bm{p}}_{s}(t) \Vert_2 > d_{\mathrm{safe}}, \text{ for } i \in \mathbb{N}_{[1, 7]},~ t \geq 0,
\end{align}
where $\hat{\bm{p}}_{s}$
is a support point on the boundary of the capsule with the minimum distance to the $i^{\text{th}}$ robot joint, as shown in Fig. \ref{fig:distance-calculation}.
We define a candidate ECBF $h_i(t, \bm{x}_i)$ as
\begin{align}
\label{eq:h}
    h_i(t, \bm{x}_i) &= \Vert \bm{x}_{i}(t) - \hat{\bm{p}}_{s}(t) \Vert_2^2 - d_{\mathrm{safe}}^2.
\end{align}
%
%
%
To modify the applied forces, $\bm{f_h}$, to guarantee safety, a quadratic programming (QP) based safety filter is employed. Considering the position constraints with relative degree 2, the optimization problem can be formulated as minimizing the normed difference between $\bm{f_h}$ and the actual input $\bm{u}_{\rm act}$, in order to maintain the robot within the safety region. In other words, $\bm{u}_{\mathrm{act}}(\bm{x}, t)$ is the control input that minimizes the following QP
\begin{subequations}
\label{eq:problemECBF}
\begin{align}
    &\minimize_{\bm{u}_\mathrm{act} \in \mathcal{U}} 
     \| \bm{f_h} - \bm{u}_{\mathrm{act}} \|^2,
\intertext{subject to}
    &L_{\bm{f}}^2 h_i(t,\bm{x}) 
    + L_{\bm{g}} L_{\bm{f}} h_i(t,\bm{x})\bm{u}_{\mathrm{act}} 
    + k_1 h_i(t,\bm{x}) 
    \notag \\
    &\qquad + k_2 L_{\bm{f}} h_i(t, \bm{x}) \geq 0, i \in \mathbb{N}_{[1, 7]},
\end{align}
\end{subequations}
where $h_i(t,\bm{x})$ is as defined in (\ref{eq:h}). The robot joint torques are then obtained as $\bm{\tau}_{\mathrm{act}} = \bm{J}^{\intercal}(\bm{q})\bm{u}_{\mathrm{act}}$.
Further details on the implementation of ECBFs are given in \cite{nguyen2016exponential}.

\subsection{System stability analysis} \label{sec:cbfclfqp}

Proving the global stability of our ECBF-NMPC dynamic motion planning framework is challenging as it incorporates approximate numerical methods, a neural network-based motion predictor, the GJK algorithm, and a safety filter. In order to guarantee local asymptotic stability, a control Lyapunov constraint is employed in the ECBF formulation and activated when the robot is close to the destination.

\noindent\textbf{Property 1 \cite{he2015neural}:} The inertia matrix $\bm{M}(\bm{q})$ and $\bm{M_x}$ are symmetric positive definite.

\noindent\textbf{Property 2 \cite{he2015neural}:} The matrix $2\bm{C}_x(\bm{q}, \dot{\bm{q}}) - \dot{\bm{M}}_x(\bm{q})$ is skew-symmetric.

\noindent\textbf{Definition 2} ($\mathcal{K}_{\infty}$-class function \cite{grandia2020nonlinear}):
 A continuous function $\beta:\mathbb{R}_{+}\rightarrow \mathbb{R}_{+}$ is a $\mathcal{K}_{\infty}$-class function (denoted as $\beta\in\mathcal{K}_{\infty}$), if
 \begin{itemize}
     \item $\beta(0) = 0$
     \item $\beta$ is increasing 
     \item $\lim_{a\rightarrow \infty} \beta(a) = \infty$.
 \end{itemize}
\noindent\textbf{Definition 3} (Control Lyapunov Functions \cite{minniti2021adaptive}):
%
Given the system (\ref{eq:convdyna}) and the composite error $\bm{z}(\bm{x},t) = \dot{\tilde{\bm{x}}} +\Lambda \tilde{\bm{x}}$, with $\Lambda>0$, consider a continuously differentiable function $V:\mathbb{R}^{3} \rightarrow \mathbb{R}_{+}$ on set $\mathcal{Z} = \{\bm{z}(\bm{x}, t)\in \mathbb{R}^3:\bm{x}\in \mathcal{C}, \forall t>0\}$, where $\mathcal{C}$ is the safe set as defined for the ECBF (Definition 1).  
Then $V$ is a control Lyapunov function (CLF) for the system in \eqref{eq:convdyna} if there exists $\beta_1$, $\beta_2$, $\beta_3 \in \mathcal{K}_{\infty}$ such that for all $\bm{z}\in \mathcal{Z}$
\begin{align}
    \beta_1(\Vert \bm{z} \Vert) \leq V(\bm{z}) \leq \beta_2(\Vert \bm{z} \Vert),
    \label{eq:clf1}
    \\
    \inf_{\bm{u}\in \mathcal{U}} \dot{V}(\bm{z}, \bm{u}) \leq -\beta_3(\Vert \bm{z} \Vert).
    \label{eq:clf2}
\end{align}
%
%
%
The existence of a CLF implies the existence of a control law that renders the closed-loop system locally asymptotically stable \cite[Thm. 2.5]{Sontag83}.
The target when employing a CLF constraint is to make the composite error $\bm{z}(\bm{x},\dot{\bm{x}})$, which models the generalized Cartesian position and velocity errors with respect to the desired references, locally asymptotically converge to 0.
Then, the problem \eqref{eq:problemECBF} needs to satisfy both the ECBF and CLF constraints, which is written as
\begin{subequations}
\label{eq:problemECBF2}
\begin{align}
     \minimize_{\bm{u}_\mathrm{act} \in \mathcal{U}} 
     &
     \| \bm{f_h} - \bm{u}_{\mathrm{act}} \|^2,
\intertext{subject to}
    &L_f^2 h_i(t,\bm{x}_i) 
    + L_g L_f h_i(t,\bm{x}_i)\bm{u}_{\mathrm{act}, i} 
    + k_1 h_i(t,\bm{x}_i) 
    \notag \\
    &\qquad + k_2 L_f h_i(t, \bm{x}_i) \geq 0, i \in \mathbb{}_{[1, 7]},\\
    &\bm{z}_i^{\intercal}\left(\bm{u}_{{\rm act}, i}+k_z\frac{\bm{z}_i}{\Vert \bm{z}_i \Vert + c_1}\right) - \bm{z}_i^{\intercal}\bm{K}_D\bm{z}_i \geq 0,
    \notag\\
    &\qquad\qquad i \in \mathbb{}_{[1, 7]},
    \label{con:CLFineq}
\end{align}
\end{subequations}
where $\bm{K}_D$ is a symmetric positive definite matrix.  
The candidate Lyapunov function $V$ is defined as
%
%


\begin{align}
\label{eq:CLFCBF}
    V(\bm{z}_i) = \tfrac{1}{2}\bm{z}_i^{\intercal}\bm{M_x}\bm{z}_i.
\end{align}
Since $\bm{M_x}(\bm{q})$ is positive definite for all $\bm{q}$, its minimum eigenvalue is positive. Assuming that $\inf_{\bm{q}}\lambda_{\min}(\bm{M_x}(\bm{q})) > 0$ and $\sup_{\bm{q}}\lambda_{\max}(\bm{M_x}(\bm{q})) < \infty$, the inequality in \eqref{eq:clf1} is satisfied with 
$\beta_1(\|z\|) = \inf_{\bm{q}}\lambda_{\min}(\bm{M_x}(\bm{q})) \|z\|^2$ and $\beta_2(\|z\|) = \sup_{\bm{q}}\lambda_{\max}(\bm{M_x}(\bm{q})) \|z\|^2$.

Differentiating \eqref{eq:CLFCBF} with respect to time and using Property 1 to combine terms, gives 
\begin{align}
\label{eq:final1}
    \dot{V} = \tfrac{1}{2}\bm{z}_i^{\intercal}\dot{\bm{M}}_{\bm{x}}\bm{z}_i + \bm{z}_i^{\intercal}\bm{M_x}\dot{\bm{z}}_i
\end{align}
Noting that $\dot{\bm{z}}_i = \ddot{\bm{x}}_i - \ddot{\bm{x}}_{d,i} +\Lambda (\dot{\bm{x}}_i - \dot{\bm{x}}_{d,i})$, and substituting for $\dot{\bm{M}}_{\bm{x}}\ddot{\bm{x}}_i$ using \eqref{eq:cartesiandy} with $\bm{f_h}$ replaced by the modified control force $\bm{f_h} - \bm{u}_{{\rm act,} i}$, $\bm{M_x}\dot{\bm{z}}_i$ can be written as
\begin{align}
\label{eq:final2}
    \bm{M_x}\dot{\bm{z}}_i &= \bm{f_h} - \bm{u}_{{\rm act,} i} - \bm{C_x}(\bm{z}_i + \dot{\bm{x}}_{d,i}
    \notag
    \\
    &\quad- \Lambda(\bm{x}_i - \bm{x}_{d,i}))- \bm{g_x} - \bm{M_x}(\ddot{\bm{x}}_{d,i} - \Lambda\dot{\tilde{\bm{x}}}_i)
\end{align}
Substituting \eqref{eq:final2} into \eqref{eq:final1} yields

 \begin{align}   
    \dot{V} &=\tfrac{1}{2}\bm{z}_i^{\intercal}\dot{\bm{M}}_{\bm{x}}\bm{z}_i + \bm{z}_i^{\intercal}[\bm{f_h} - \bm{u}_{{\rm act}, i} - \bm{C_x}(\bm{z}_i + \dot{\bm{x}}_{d,i}
    \notag
    \\
    &\quad- \Lambda(\bm{x}_i - \bm{x}_{d,i}))- \bm{g_x} - \bm{M_x}(\ddot{\bm{x}}_{d,i} - \Lambda\dot{\tilde{\bm{x}}}_i)]
    \notag
    \\
    &=\tfrac{1}{2}\bm{z}_i^{\intercal}(\dot{\bm{M_x}}-2\bm{C_x})\bm{z}_i + \bm{z}_i^{\intercal}[\bm{f_h} - \bm{u}_{{\rm act}, i}
    \notag
    \\
    &\quad- \bm{C_x}(\dot{\bm{x}}_{d,i} - \Lambda\tilde{\bm{x}}_{i})- \bm{g_x}] - \bm{z}_i^{\intercal}\bm{M_x}(\ddot{\bm{x}}_{d,i} - \Lambda\dot{\tilde{\bm{x}}}_i).
    \label{eq:dotlyapunov1}
\end{align}
By Property 2, $\bm{z}_i^{\intercal}(\dot{\bm{M_x}}-2\bm{C_x})\bm{z}_i=0$, hence the derivative of $V$ can be expressed as
\begin{multline}
     \dot{V}
     {}={}
     \bm{z}_i^{\intercal}[(\bm{f_h} - \bm{u}_{{\rm act}, i}) - \bm{C_x}(\dot{\bm{x}}_{d,i} - \Lambda\tilde{\bm{x}}_i) - \bm{g_x}]
     \\ 
     {}-{} \bm{z}_i^{\intercal}\bm{M_x}(\ddot{\bm{x}}_{d,i} - \Lambda\dot{\tilde{\bm{x}}}_i).
\label{eq:dotlyapunov2}
\end{multline}

Substituting \eqref{eq:controlforcenew} into \eqref{eq:dotlyapunov2} yields
\begin{align}
    \dot{V} =  - \bm{z}_i^{\intercal}\bm{u}_{{\rm act}, i} -\frac{k_z\bm{z}_i^{\intercal}\bm{z}_i}{\Vert \bm{z}_i \Vert + c_1}.
\end{align}
According to Definition 2 and \eqref{con:CLFineq}, the stability of the low-level controller is guaranteed when $\dot{V} \leq -\bm{z}_i^{\intercal}\bm{K}_D\bm{z}_i$ assuming problem \eqref{eq:problemECBF2} is feasible.

\section{Experimental Setup}\label{sec:ex}
\begin{figure}
\centering
\includegraphics[width=0.8\columnwidth]{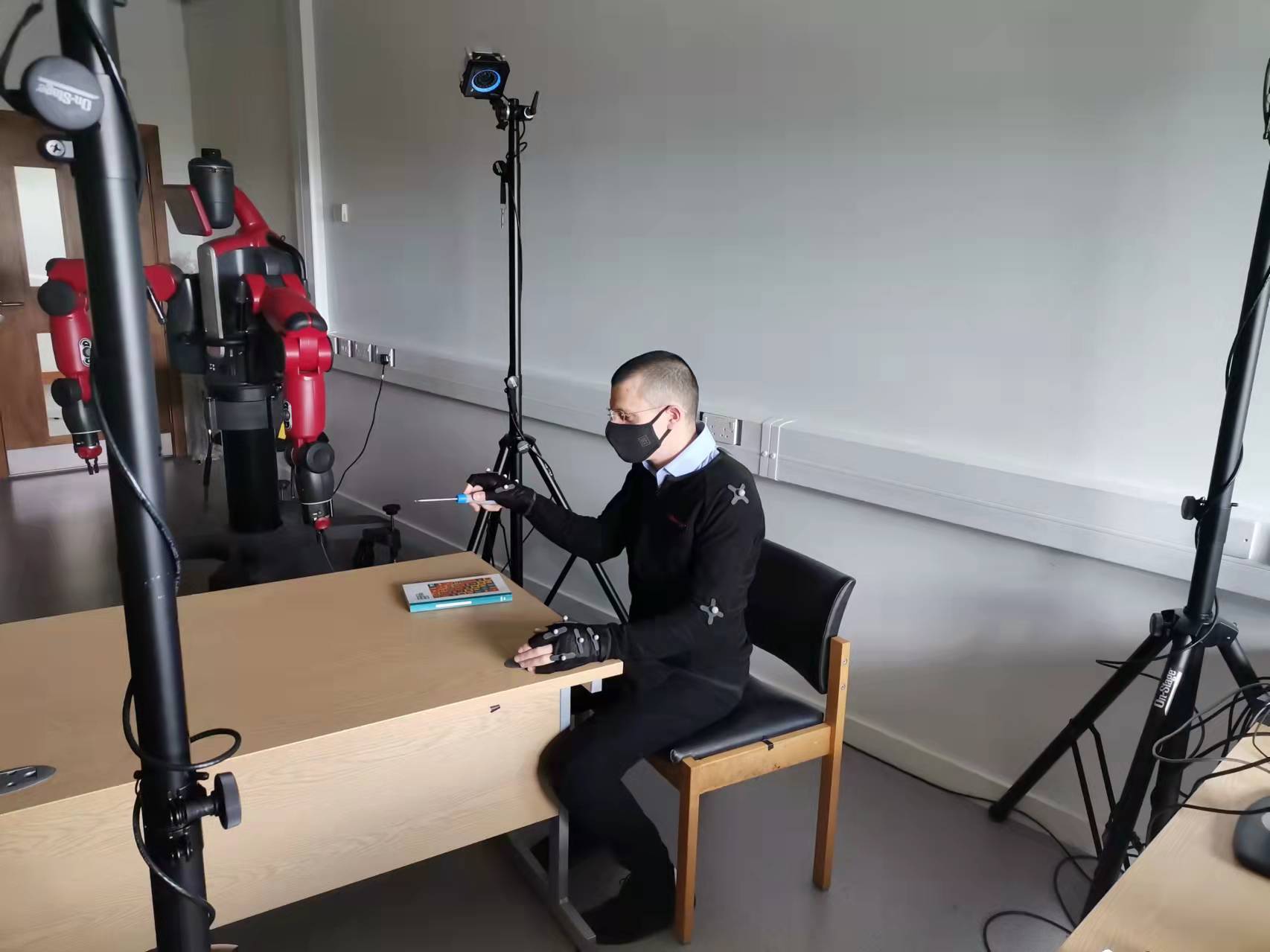}
\label{fig:capture}
\end{figure}

The collaboration scenario shown in Fig. \ref{fig:capture} is a typical screw-driver usage scenario where the robot holding the screw at an initial position, delivers it to the human operator before the operator picks the screwdriver up. The operator's responsibility is to use the screwdriver and take the screw from the robot end-effector, which stops at a safe interactive distance $d_{\mathrm{safe}}$. Active collision avoidance is activated when the human body appears in the robot's original trajectory. For seamless human-robot collaboration, the robot starts to calculate a new trajectory based on the predicted future pose sequence for the operator (1-sec horizon, $N_h$ = 20). The overall procedure for the screw-driver usage task is illustrated in Algorithm 1. After the initialization of the robot and the OptiTrack camera system, the software enters the main execution loop. Here, the action being undertaken by the operator is predicted at each sample instant. If an action requiring interaction is predicted the software calls Algorithm 2 to compute the required robot joint torques in accordance with the proposed ECBF-NMPC dynamic motion planning framework.

\begin{algorithm}
\DontPrintSemicolon
  
  \KwInput{$f_{\mathrm{recg}}$ (pretrained task classifier).}
  \KwData{OptiTrack data stream }
  Initialize the robot to its original position.\;
  Initialize the Optitrack camera system and $\bm{P}_{\mathrm{prev}}$.\;
   \For{$t = 1,\dots$}{
   Read $\bm{P}(t)$ from the OptiTrack camera system.\;
   $\bm{P}_{\mathrm{prev}}(t) = [\bm{P}(t - \tau_o + 1),\dots,\bm{P}(t)]$.\;
   $\hat{y} = f_{\mathrm{recg}}(\bm{P}_{\mathrm{prev}}(t))$.\;
  \eIf{$\hat{y}(t) \in \mathcal{A}_{\rm int}$}{
        Get the current state of the robot $\bm{x}$.\;
        $\hat{\bm{\tau}}_{\mathrm{act}}(t) = Algorithm2(\bm{P}_{\mathrm{prev}}(t), \bm{x})$.\;
        Apply $\hat{\bm{\tau}}_{\mathrm{act}}(t)$ to the robot.\;
        }
    { 
    Stop the robot.\;
    }}
\caption{Vision-based screw-driver usage task execution}
\end{algorithm}

\begin{algorithm}
\DontPrintSemicolon
  
  \KwInput{$f_{\mathrm{pred}}$ (pretrained motion predictor), $\bm{P}_{\mathrm{prev}}$ (previous one second motion sequence), $\bm{x}$ (current state of the robot).}
  \KwOutput{input to robot $\bm{\tau}_{\mathrm{act}}$}
  Define the controller.\;
        $[\hat{\bm{p}}_{\mathrm{rh}}, \hat{\bm{p}}_{\mathrm{o}}] = f_{\mathrm{pred}}(\bm{P}_{\mathrm{prev}})$\;
        Build the capsule-format representation of the human body and compute the minimum distance $\lambda(\bm{x}, \hat{\bm{p}}_{\bm{o}}))$.\;
        Solve the defined problem in OpEn and compute the optimal input to the high-level controller $\hat{\bm{u}}^{\star}$.\;
        Compute the desired robot state $\bm{q}_d$ and $\dot{\bm{q}}_d$.\; 
        Design the control input to the safety filter $\bm{f_h}(t)$.\;
        Solve the ECBF quadratic program to determine $\bm{u}_{act}(t)$.\;
        Compute the input to the robot: $\bm{\tau}_{\mathrm{act}}(t) = \bm{J}^{\intercal}(\bm{q}(t))\bm{u}_{\mathrm{act}}(t)$.\;
\caption{Motion-prediction-based path planning}
\end{algorithm}
\subsection{Motion Capture}
In this work, we generate high-quality skeleton data using an OptiTrack camera system. The volunteer performs the required behavior in a working environment with three cameras, with the volunteer's behavior tracked from different perspectives, as shown in Fig 5(a). The human skeleton model generated has 32 joints, with joint data recorded at a sampling interval of 50 ms (i.e., 20 frames per second).

The capture space evaluated was based on typical use and had an extent of $2.5\text{m} \times 3.5\text{m} \times 2\text{m} = 17.5\text{m}^3$. This space can be covered by 3 cameras. The number of cameras needed will increase for larger spaces (volumes). We have opted for a marker-based approach in this work, regardless of the high cost of system setup and calibration, to ensure high-quality input into the perception system. 

\subsection{Vision System}
For the vision module, we use a non-local graph convolution (NGC) based method developed in  \cite{zhang2021non} for action recognition and motion prediction. This implements a spatial-temporal attention mechanism to extract nonlocal spatial-temporal features enabling more accurate motion prediction over longer horizons than alternative approaches. As this model, referred to as the NGC-model, performs well for the Human 3.6M, CMU Mocap and NTU RGB-D datasets (see \cite{zhang2021non}), we repurpose it to predict the human motion associated with the screw-driver task and classify it into one of 6 subtasks that constitute the overall activity, namely, 'pick up', 'move forward', 'take the screw', 'operate screw-driver', 'move backward' and 'put down'. This is achieved by retraining the network with a dataset collected for the screw-driver task consisting of skeletal data and associated subtask labels for three repetitions of the task. In this work, 4372 frames are used for training, while 1287 frames are used for testing. Each frame includes 3D joint positions with 32 joints, as well as ground-truth action labels for each frame.

 \begin{figure}
\centering
\includegraphics[width=6cm,height = 4.5cm]{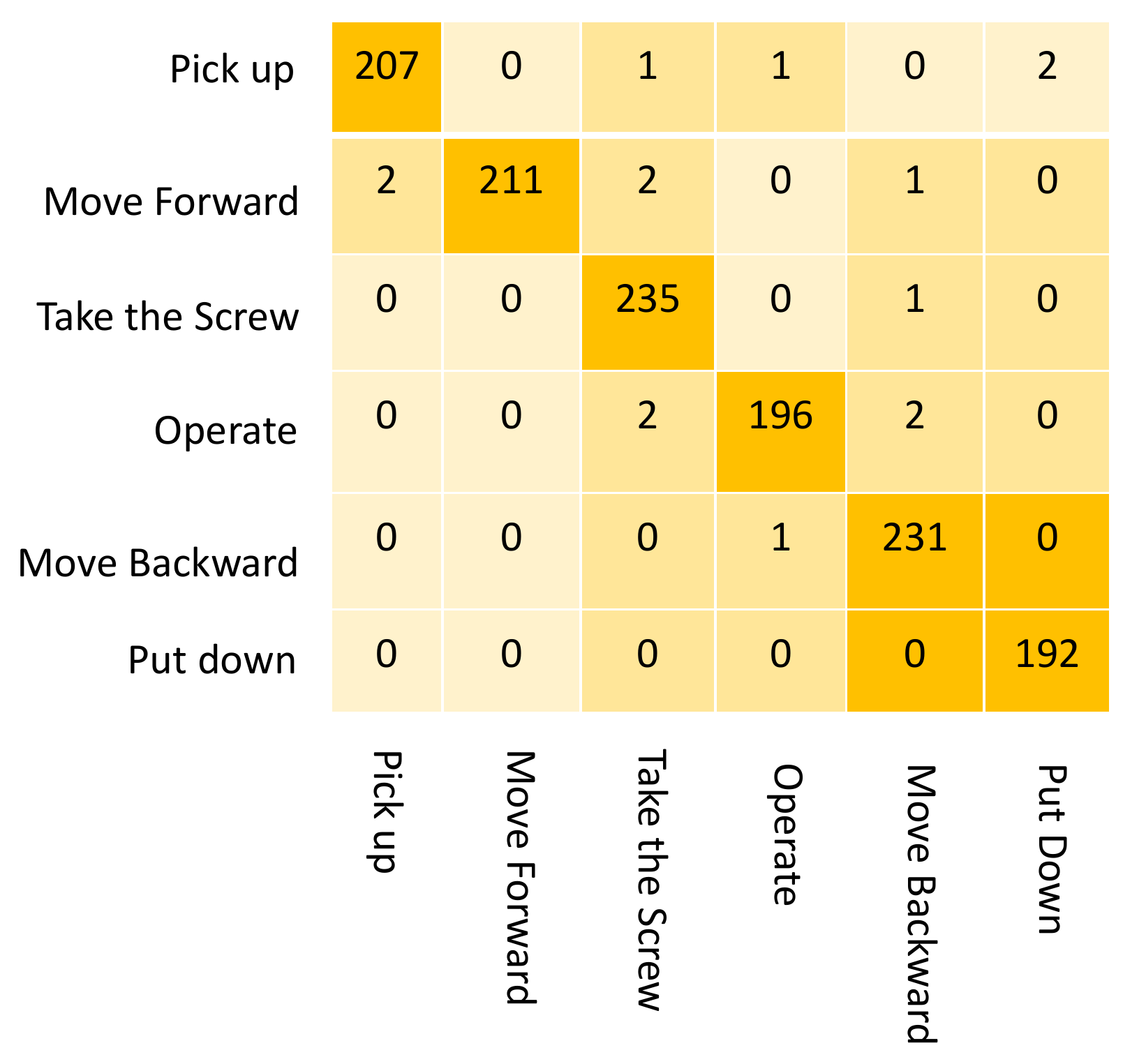}
\caption{Confusion matrix describing the accuracy of action recognition in the screw-driver usage task.}
\label{fig:visionMAE}
\end{figure}

\begin{table}[htbp]
  \centering
  \caption{Comparison of MPJPE between models for 1000ms prediction.}
  \resizebox{\columnwidth}{!}{%
    \begin{tabular}{|c|c|c|c|c|c|c|}
    \hline
          & pick up & Move forward & Take the screw & Operate & Move backward & Put Down \\
    \hline
    LSTM-3LR \cite{gui2018teaching} & 2.41  & 2.26  & 2.73  & 2.52  & 2.68  & 2.88 \\
    \hline
    Res-sup \cite{martinez2017human} & 2.18  & 2.14  & 2.42  & 2.38  & 2.57  & 2.71 \\
    \hline
    Traj-GCN \cite{mao2019learning} & 1.95  & 2.01  & 2.29  & 2.21  & 2.46  & 2.61 \\
    \hline
    SPGSN \cite{10.1007/978-3-031-20068-7_2} & 1.33  & 1.96  & 2.01  & 2.09  & 2.41  & 2.37 \\
    \hline
    \textbf{Ours} & \textbf{1.29} & \textbf{1.92} & \textbf{1.98} & \textbf{2.07} & \textbf{2.39} & \textbf{2.35} \\
    \hline
    \end{tabular}%
    }
  \label{tab:1}%
\end{table}%

\subsubsection{Performance of the Vision Module} 
 As shown by the confusion matrix in Fig. \ref{fig:visionMAE}(a), the model has a high prediction accuracy for each subtask (99.0\%, 100\%, 97.9\%, 99.0\%, 98.3\%, and 99.5\%) and achieves an overall accuracy of 98.9\%. For motion prediction, the mean per joint position error (MPJPE) is used to quantify the mean deviation between the predicted and actual locations of the human body joints in each frame. As shown in Table.\ref{tab:1}, the NGC-model achieves the lowest error compared to selected baselines models from the literature (LSTM-3LR \cite{gui2018teaching}, Res-sup \cite{martinez2017human}, Traj-GCN \cite{mao2019learning} and SPGSN \cite{10.1007/978-3-031-20068-7_2}). Table.\ref{tab:1} compares the MPJPE of our model and the baselines for all actions, with the prediction horizon of 1000ms. It shows that our model obtains the best performance for all the actions. 
 
 This shows that the proposed NGC model can accurately and effectively identify the intended human body movement in real-time. Using this capability the HRC system can understand the intention of the human being when the human body is beginning an action, making it feasible for it to provide assistance to the human to complete the task efficiently.
\begin{figure*}
\centering
\subfloat[]{\includegraphics[width=6cm, height = 4.2cm]{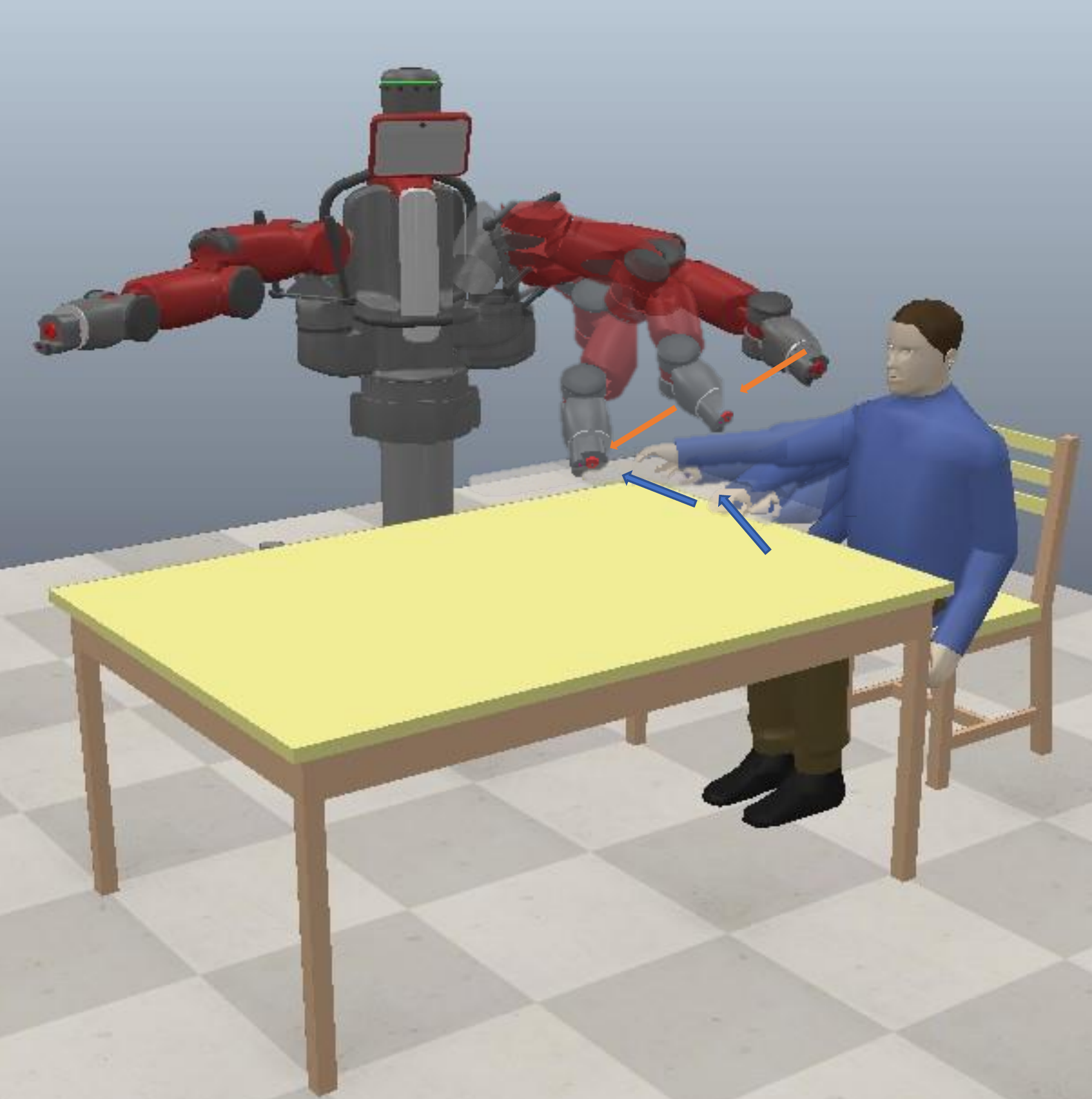}}\hfil
\subfloat[]{\includegraphics[width=6cm, height = 4.2cm]{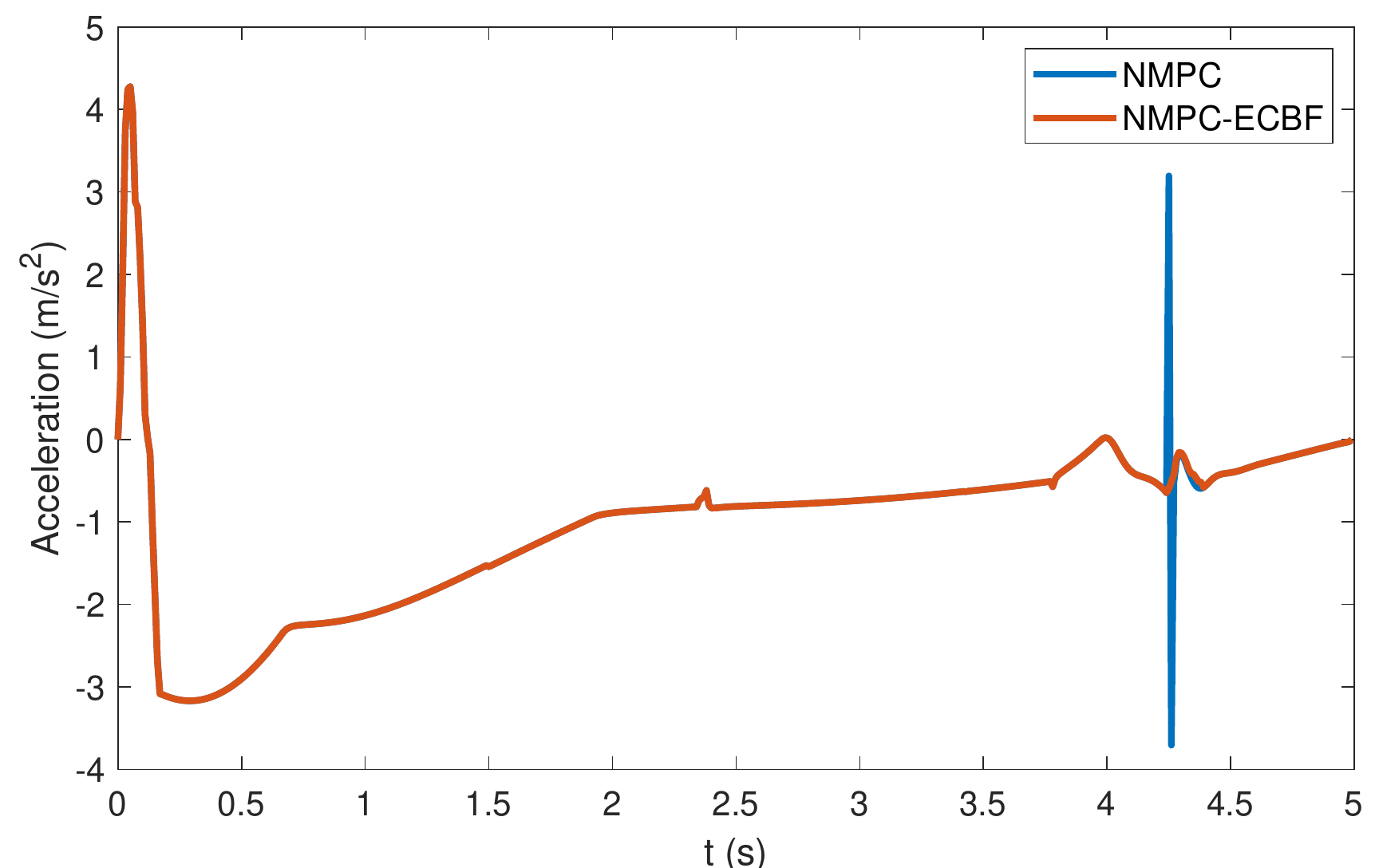}}\hfil 
\subfloat[]{\includegraphics[width=6cm, height = 4.2cm]{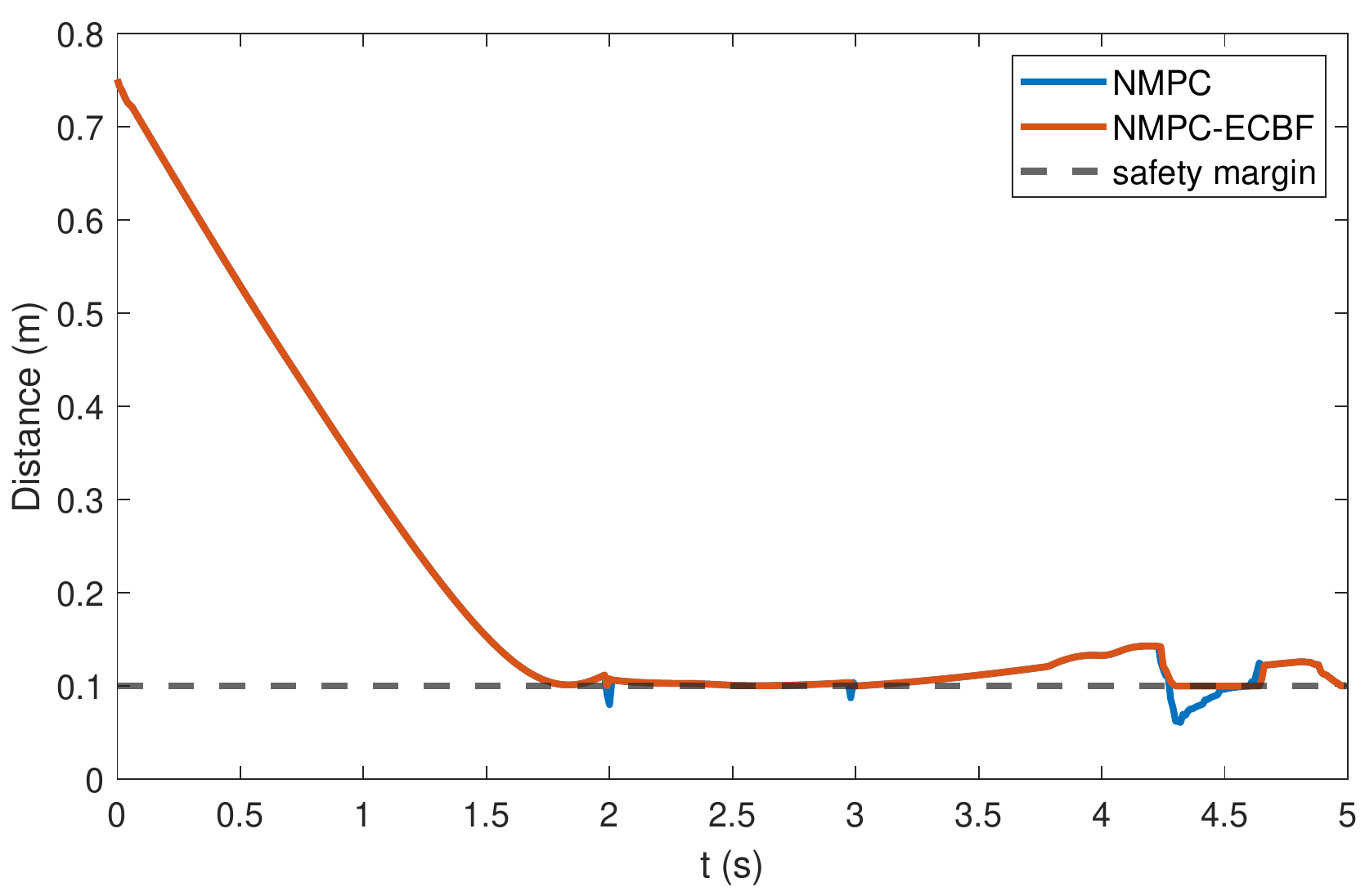}}
\caption{(a) HRC scenario when the initial position is close to the operator; (b) The acceleration of the end-effector; (c) The minimum distance to the obstacles among all the robot links}.
\label{fig:scenario1_config}
\end{figure*}

\subsection{Control System Performance Test}
To evaluate NMPC-ECBF we implement it for a 7-DOF Baxter robot simulated in CoppeliaSim using a standard NMPC controller as a baseline comparison \cite{oleinikov2021safety}.  The system model employed in the MPC controller is discretized using the Euler method. The weight matrices in $\ell$ and $\ell_f$ are selected as $\bm{Q}_v = \mathrm{diag}$(1,1,1,1,1,1,1), $\bm{Q}_p = \mathrm{diag}$(5,5,5,1,1,0), $\bm{R}_v =\mathrm{diag}$(0.1,0.1,0.1,0.1,0.1,0.1,0.1) and $\bm{R}_p = \mathrm{diag}$(3,3,3,0,0,0). Solution tolerances $\epsilon_a = 10^{- 2}$ and $\epsilon_b = 10^{- 4}$ are set for the fixed-point residual and the infeasibility, respectively. For the safety filter, safe set $\mathcal{U}$ for input $\bm{u}_{\mathrm{act}}$ is $(-\unit[40]{N}, \unit[40]{N})$. The parameters of the low-level controller are selected as $k_z = 5$ and $c_1 = 0.01$. Noting the maximum joint prediction errors in Fig. \ref{fig:visionMAE} (c), we set the safe distance ($d_{\rm safe}$) as $10$ cm. The ECBF coefficients $\bm{k}_b$ are chosen as $[7 ,7]$, while $\bm{K}_D = \mathrm{diag}$(5,5,5).

In the CoppeliaSim simulation, the movement of the human is modeled using the recorded trajectories of the skeleton joints. The Baxter robot starts from different initial positions and is required to reach the estimated position without collision with the simulated human, as shown in Fig. \ref{fig:scenario1_config} (a). 
The acceleration of the robot end-effector and the minimum distance between the robot and human obstacle during its motion in scenario 1 is plotted in Fig. \ref{fig:scenario1_config} (b) and (c), respectively. The acceleration plot shows that NMPC-ECBF yields fewer abrupt changes in acceleration than the baseline NMPC path planner. The minimum distance plot highlights that the trajectory generated by NMPC violates the safe area at several points during the robot motion and that the inclusion of the ECBF safety filter has the desired effect of maintaining the motion within the safe area.

To further assess the robustness and reliability of the controller we perform 100 dynamic path planning HRC experiments with each one corresponding to a randomly generated robot start position and one of the 3 sets of human screw-driver usage task movement trajectories. For each experiment, we recorded the maximum acceleration of the end-effector and the minimum distance between the end-effector and the target safety margin during the task execution. Table \ref{tab:100tests} reports the mean and most extreme values observed over the 100 experiments. The results show that NMCP-ECBF is able to strictly keep the robot motion within the safe area for all 100 simulations, with no violations of the safety margin ($d=0$). In contrast, the NMPC path planner frequently breaches the safety margin, with an average violation of $1.62$ cm and a maximum violation of $3.26$ cm. Furthermore, NMPC-ECBF outperforms NMPC in terms of the average and maximum joint accelerations, achieving a reduction in these values of $48.0\%$ and $78.2\%$, respectively.

\begin{table}[htbp]
  \centering
  \caption{Controller performance for 100 HRC task instances.}
  \small\addtolength{\tabcolsep}{-5pt}
  \resizebox{\linewidth}{!}{%
    \begin{tabular}{|c|c|c|c|c|}
    \hline
          & max acc (m/s$^2$) & min d (cm) & avg max acc (m/s$^2$) & avg min d (cm) \\
    \hline
    NMPC  & 12.41  & -3.26 & 2.73  & -1.62 \\
    \hline
    NMPC-ECBF & 2.70  & 0     & 1.42  & 0 \\
    \hline
    \end{tabular}}%
  \label{tab:100tests}%
\end{table}%

\begin{figure}
\centering
\includegraphics[width=\columnwidth]{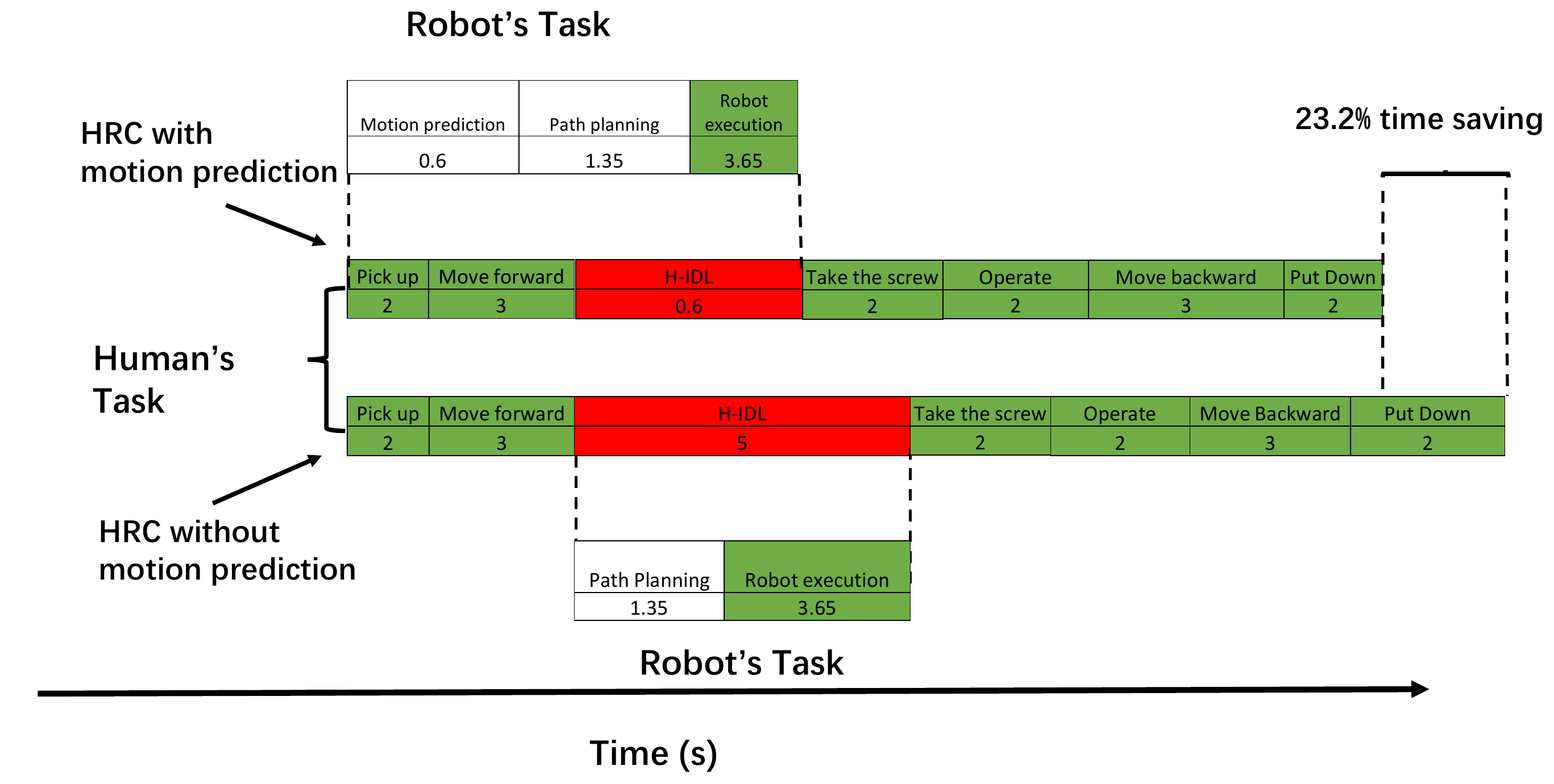}
\caption{Task assignment chart for the baseline approach of strict turn-taking, where each action is immediately followed by the next action of the other teammate.}
\label{fig12}
\end{figure}

\subsection{Idle time for agents in the HRC scenario}
Fig. \ref{fig12} compares the human idle time (H-IDL) and robot idle time (R-IDL) in the screw-driver usage HRC task (i.e. Scenario 1) to highlight the benefit of integrating motion prediction into HRC. The robot is triggered when the human's action is detected to be `pick up' and is stopped when the `take the screw' action is detected. In the proposed screw-driver usage task, the original length of this HRC task is 14 seconds, which is the time used in human-human collaboration.
Red blocks indicate when the operator is in an idle state, waiting for the screw to arrive. White blocks are the times the robot is idle for computation and data loading. Green blocks are the times when both agents are executing the commanded task. The R-IDL when employing motion-prediction-based control consists of the time for motion prediction, and path planning. The robot execution time takes the majority of the total time when employing the motion predictor and embedded optimization engine. In the HRC task implemented without motion prediction, the robot has to start path planning after the human operator moves their right hand to the target interactive position. 

Table. \ref{tab:efficiency} compares the human idle time (H-IDL) and robot idle time (R-IDL) for the human-robot collaboration task without motion prediction. In this work, the HRC without motion prediction can be treated as a strict turn-taking task, where each agent strictly takes action after the partner finishes. Using the motion-prediction-based approach, we can achieve a H-IDL rate of 4.1\% and a R-IDL rate of 13.4\% in the HRC task. Without motion prediction, the H-IDL rate and R-IDL rate are 26.3\% and 7.1\%, respectively. Therefore, using motion prediction reduces the idle time for both the human and the robot in the HRC task. As a result, the total execution time is reduced from 19.0  to 14.6 seconds, a saving of 23.2\%. 

\begin{table}[htbp]
  \centering
  \caption{Comparison of the idle time between the tasks with or without motion prediction.}
  \resizebox{\columnwidth}{!}{%
    \begin{tabular}{|c|c|c|}
    \hline
          & With Motion Prediction (s) & Without Motion Prediction (s) \\
    \hline
    H-IDL & 0.6  & 5 \\
    \hline
    R-IDL & 1.95  & 1.35  \\
    \hline
    Total Time & 14.6 & 19.0 \\
    \hline
\end{tabular}}%
\label{tab:efficiency}%
\end{table}%

\section{Conclusions} \label{sec:con}
 The proposed NMPC-ECBF control framework allows a robot to safely replan its motion in the presence of a human, delivering smooth acceleration and adaptivity in the presence of uncertainty. The results demonstrate the necessity of adding a safety filter into the NMPC controller in the trajectory planning problem. This represents a promising approach for manufacturing applications due to its characteristic of providing a strict constraint on robot motion. In addition, by employing human motion prediction within the framework tasks can be executed more efficiently with a lower idle rate achievable for both agents in a HRC task than with traditional turn-taking task execution.
 
Future work will be devoted to testing the proposed algorithm when there is uncertainty regarding the system dynamics (e.g., caused by joint friction). The possibility of promoting the proposed model to scenarios with multiple robots and coworkers will also be investigated. In addition, we will also explore the applicability of our approach to scenarios involving both physical and non-physical HRC interactions, with the robot switching between different control modes based on the prevailing circumstances, as determined by a higher-level supervisory system. What’s more, we will integrate multimodal learning (e.g. radar sensors, tactile sensors, and auditory sensors) into the HRC system for more robust and safer collaboration. 
 
%
%
%
%
%

\printbibliography

\end{document}